\documentclass[10pt,twocolumn,letterpaper]{article}

\usepackage[pagenumbers]{cvpr} 

\usepackage{graphicx}
\usepackage{amsmath}
\usepackage{amssymb}
\usepackage{booktabs}

\usepackage{epsfig}
\usepackage{graphicx}
\usepackage{tabularx,booktabs}
\usepackage{multirow}
\usepackage{wrapfig}
\newcolumntype{C}{>{\centering\arraybackslash}X} 
\newcolumntype{S}{>{\small}c}
\usepackage{adjustbox}
\usepackage{overpic}
\usepackage[dvipsnames]{xcolor}

\usepackage[pagebackref,breaklinks,colorlinks]{hyperref}

\usepackage[capitalize]{cleveref}
\crefname{section}{Sec.}{Secs.}
\Crefname{section}{Section}{Sections}
\Crefname{table}{Table}{Tables}
\crefname{table}{Tab.}{Tabs.}

\usepackage{xcolor}
\definecolor{light-gray}{gray}{0.95}

\newcommand{\nomod}[1]{{#1}$\text{-X-NK}^\dagger$}
\newcommand{\modified}[1]{{#1}$\text{-M12-NK}^\dagger$}

\newcommand{\appx}{{\raise.17ex\hbox{$\scriptstyle\sim$}}}

\newcommand{\zoedmnk}{\textit{ZoeD-M12-NK}}

\usepackage{pifont}
\newcommand{\cmark}{\ding{51}}%
\newcommand{\xmark}{\ding{55}}%

\def\cvprPaperID{1832} 
\def\confName{CVPR}
\def\confYear{2023}

\begin{document}

\title{{\textcolor{teal}{ZoeDepth}}: {\textcolor{teal}{Z}}ero-sh{\textcolor{teal}{o}}t Transf{\textcolor{teal}{e}}r by Combining Relative and Metric {\textcolor{teal}{Depth}}}

\author{Shariq Farooq Bhat\\
KAUST
\and
Reiner Birkl\\
Intel
\and
Diana Wofk\\
Intel
\and
Peter Wonka\\
KAUST
\and
Matthias M{\"u}ller\\
Intel
}

\twocolumn[{%
\renewcommand\twocolumn[1][]{#1}%
\maketitle
\begin{center}
    \centering
    \captionsetup{type=figure}
    \includegraphics[width=\textwidth,height=4cm]{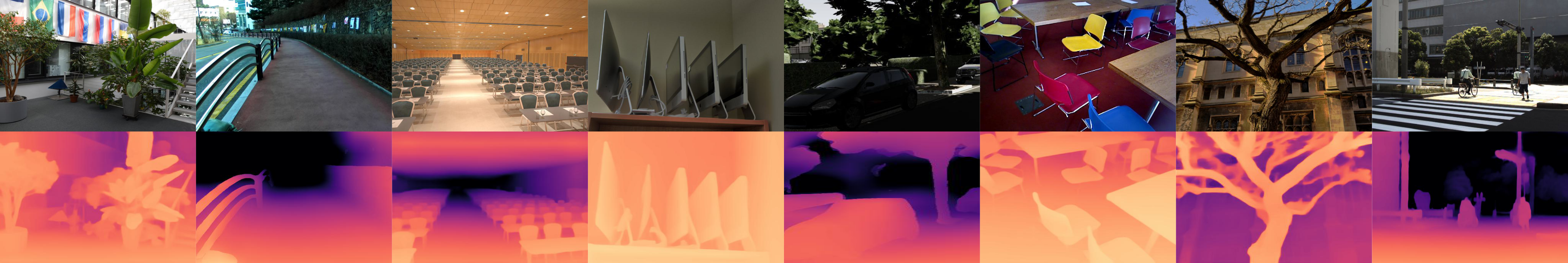}
    \captionof{figure}{\textbf{Zero-shot transfer.} Our single multi-domain metric depth estimation model can be applied across domains, indoor or outdoor, simulated or real. \textbf{Top:}~Input RGB. \textbf{Bottom:} Predicted depth. \textbf{From left to right: }iBims-1, DIML Outdoor, Hypersim, DIODE Indoor, vKITTI2, SUN-RGBD, DIODE Outdoor and DDAD.}
    \label{fig:intro}
\end{center}%
}]


\begin{abstract}
This paper tackles the problem of depth estimation from a single image. Existing work either focuses on generalization performance disregarding metric scale, \ie relative depth estimation, or state-of-the-art results on specific datasets, \ie metric depth estimation. We propose the first approach that combines both worlds, leading to a model with excellent generalization performance while maintaining metric scale. Our flagship model, ZoeD-M12-NK, is pre-trained on 12 datasets using relative depth and fine-tuned on two datasets using metric depth. We use a lightweight head with a novel bin adjustment design called metric bins module for each domain. During inference, each input image is automatically routed to the appropriate head using a latent classifier.
Our framework admits multiple configurations depending on the datasets used for relative depth pre-training and metric fine-tuning. Without pre-training, we can already significantly improve the state of the art (SOTA) on the NYU Depth v2 indoor dataset. Pre-training on twelve datasets and fine-tuning on the NYU Depth v2 indoor dataset, we can further improve SOTA for a total of 21\% in terms of relative absolute error (REL). Finally, ZoeD-M12-NK is the first model that can jointly train on multiple datasets (NYU Depth v2 and KITTI) without a significant drop in performance and achieve unprecedented zero-shot generalization performance to eight unseen datasets from both indoor and outdoor domains. The code and pre-trained models are publicly available at \url{https://github.com/isl-org/ZoeDepth}.

\end{abstract}

\begin{figure*}
    \centering
    \includegraphics[width=2.09\columnwidth]{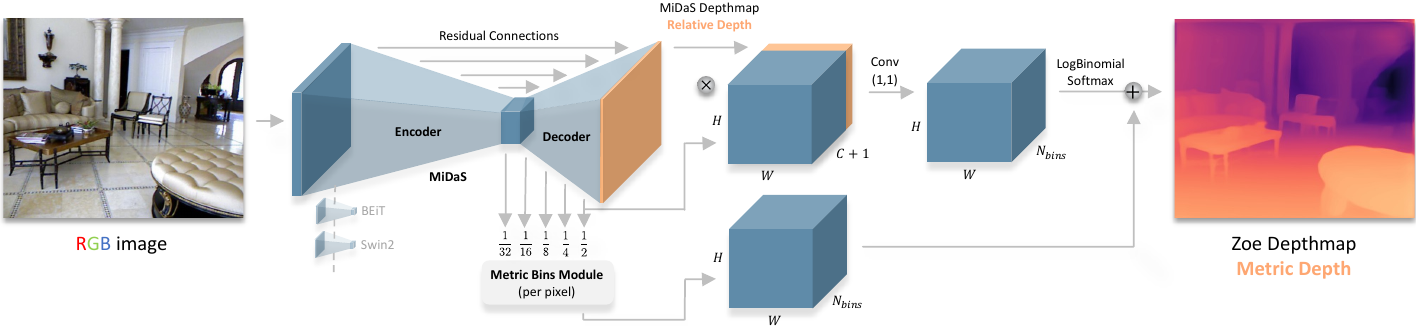}
    \caption{\textbf{ZoeDepth architecture.} An RGB image is fed into the MiDaS depth estimation framework~\cite{Ranftl2020MiDaS}. The bottleneck and succeeding four hierarchy levels of the MiDaS decoder (at $1/32$, $1/16$, $1/8$, $1/4$ and $1/2$ of the MiDaS in- and output resolution) are hooked into the metric bins module (see Fig.~\ref{fig:architecture-bin-module}). The metric bins module computes the per-pixel depth bin centers that are linearly combined to output the metric depth.
    Different transformer backbones can be utilized for the MiDaS encoder; a state-of-the-art example is BEiT\textsubscript{384}-L \cite{DBLP:journals/corr/abs-2106-08254}.
    }
    \label{fig:architecture-zoe}
\end{figure*}

\setlength{\abovedisplayskip}{3pt}
\setlength{\belowdisplayskip}{3pt}

\section{Introduction}
\label{sec:intro}


Single-image depth estimation (SIDE) is a classic problem in computer vision with many recent contributions. There are two branches of work: metric depth estimation (MDE) and relative depth estimation (RDE). The dominant branch is MDE~\cite{bhat2021adabins,bhat2022localbins, yuan2022new, li2022binsformer, SIDEreviewMERTAN2022103441}, where the goal is to estimate depth in absolute physical units, i.e. meters. The advantage of predicting metric depth is the practical utility for many downstream applications in computer vision and robotics, such as mapping, planning, navigation, object recognition, 3D reconstruction, and image editing.
However, training a single metric depth estimation model across multiple datasets often deteriorates the performance, especially when the collection includes images with large differences in depth scale, e.g. indoor and outdoor images. As a result, current MDE models usually overfit to specific datasets and do not generalize well to other datasets.

The second branch of work, relative depth estimation~\cite{SIDEreviewMERTAN2022103441, Ranftl2020MiDaS}, deals with the large depth scale variation in multiple types of environments by factoring out the scale. As a result, disparity is sufficient for supervision; metric scale and camera parameters are not required and do not need to be consistent across datasets. In RDE, depth predictions per pixel are only consistent relative to each other across image frames and the scale factor is unknown. This allows methods to be trained on a diverse set of scenes and datasets, even 3D movies \cite{Ranftl2020MiDaS}, enabling model generalizability across domains. The trade-off is that the predicted depth has no metric meaning, limiting the applications.

In this paper, we propose a two-stage framework that combines the two approaches (see Fig.~\ref{fig:architecture-zoe}). In the first stage, we train a common encoder-decoder architecture for relative depth estimation using the standard training scheme~\cite{Ranftl_2021_ICCV_DPT}. Our model first learns from a large variety of datasets in pre-training which leads to good generalization. In the second stage, we add heads for metric depth estimation to the encoder-decoder architecture and fine-tune them on metric depth datasets, using one light-weight metric head per domain (a metric had has less than 1\% of the parameters of the backbone).
During inference, an image is automatically routed to the appropriate head using a classifier on encoder features.
Adding these domain-specific heads helps the model learn metric depth while benefiting from the relative depth pre-training.
Our metric head design (dubbed metric bins module) is inspired by a recently introduced method for metric depth estimation~\cite{bhat2022localbins} that estimates a set of depth values instead of a single depth value per pixel. Similarly, we estimate a set of depth values (bins) and subsequently transform this estimation at each layer of the decoder using a novel concept we call \textit{attractors}.

Our framework is flexible and can be used in multiple different configurations. We specifically want to highlight three configurations, that improve the state-of-the-art (SOTA) in different categories of metric SIDE. These three configurations are the main contributions of this paper.

\textbf{Metric SIDE}. Without any relative pre-training, our model \textit{ZoeD-X-N} is trained only on NYU Depth v2~\cite{Silberman2012}. This configuration validates the design of our metric bins module and demonstrates that it can already improve upon the current SOTA NeWCRFs~\cite{yuan2022new} by 13.7\% on indoor depth estimation without relative depth pre-training.

\textbf{Metric SIDE with relative depth pre-training.} By conducting relative depth pre-training on 12 datasets and then conducting metric fine tuning on NYU Depth v2, our model \textit{ZoeD-M12-N} can further improve on \textit{ZoeD-X-N} by 8.5\%, leading to 21\% improvement over current published SOTA. Existing architectures do not have an established way to benefit from relative depth pre-training at a competitive level.

\textbf{Universal Metric SIDE with automatic routing.} We make a step towards universal depth estimation in the wild. Our flagship architecture \textit{ZoeD-M12-NK} uses relative pre-training on 12 datasets combined with metric fine-tuning on indoor and outdoor datasets, \ie NYU Depth v2 and KITTI, jointly. 
We evaluate this setup by first showing that it significantly outperforms SOTA on datasets it was trained on (NYU and KITTI) when compared to other models that are also trained on these two datasets jointly; we achieve an overall improvement in absolute relative error (REL) of 24.3\%. Second, our setup outperforms SOTA on 7 metric datasets it was not trained on, with up to 976.4\% ($\sim$11x) improvement in metrics; this demonstrates its unprecedented zero-shot capabilities.

\section{Related Work}
\label{sec:related}
\subsection{Single-Image Depth Estimation (SIDE)}
Supervised single-image depth estimation methods can be categorized into regressing metric depth~\cite{SIDEreviewMERTAN2022103441, jun2022depth, bhat2021adabins,bhat2022localbins, yuan2022new, li2022binsformer} and relative depth~\cite{lee2019monocular, Ranftl2020MiDaS, Ranftl_2021_ICCV_DPT, SIDEreviewMERTAN2022103441}. Metric depth models are typically trained on singular datasets, are more prone to overfitting, and typically generalize poorly to unseen environments or across varying depth ranges. Relative depth models tend to generalize better as they can be trained on more diverse datasets with relative depth annotations using scale-invariant losses. Yet, their utility for downstream tasks requiring metric depth is limited, as relative depth models regress depth with unknown scale and shift. Recent works have sought to resolve metric information in regressed depth. For example, Yin~\etal~\cite{yin2021learning} recover 3D scene shape from a single image via a two-stage framework combining monocular depth estimation with 3D point cloud encoders that are trained to predict missing depth shift and focal length. Jun~\etal~\cite{jun2022depth} decompose metric depth into normalized depth and scale features and propose a multi-decoder network where a metric depth decoder leverages relative depth features from the gradient and normalized depth decoders.
Universal depth prediction has also been investigated by the Robust Vision Challenge~\footnote{http://www.robustvision.net/} that includes indoor and outdoor domains. 
A popular idea is to discretize the target depth interval and reformulate the continuous depth regression as a classification task~\cite{Fu2018DeepOR,DABC_10.1007/978-3-030-20870-7_41,LI2018328CSWS,DSSIDERen2019DeepRS}. Ren~\etal~\cite{DSSIDERen2019DeepRS} propose a two-stage framework: first involving training a classifier to distinguish low-depth-range and high-depth-range images. Two separate networks are then trained for the respective depth ranges. We compare to the best publicly available model based on DORN~\cite{Fu2018DeepOR}.
\subsection{Distribution learning for metric SIDE}
Many conventional learning-based monocular depth estimation methods adopt encoder-decoder architectures with convolutional layers, and more recently, transformer blocks. Depth estimation is commonly treated as a per-pixel regression task. An evolving line of work seeks to reformulate depth estimation as a combined classification-regression problem that reasons about distributions of depth values across an image. AdaBins \cite{bhat2021adabins} extends standard encoder-decoder backbones with a transformer-based module that discretizes predicted depth ranges into bins, where bin widths are determined adaptively per image; the final depth estimation is computed as a linear combination of bin centers. LocalBins \cite{bhat2022localbins} builds on this concept by considering depth distributions within local neighborhoods of a given pixel instead of globally over the image, as well as computing bin embeddings in a multi-scale fashion across decoder layers. PixelBins~\cite{pixelbinsSarwari:EECS-2021-32} simplifies AdaBins by replacing transformer block with convolutions, reducing complexity. BinsFormer \cite{li2022binsformer} incorporates an auxiliary scene classification query to guide bin generation and also utilizes a multi-scale strategy to refine adaptively-generated bins. PixelFormer \cite{agarwal2022attention} treats depth estimation as pixel queries that are refined via skip attention and that are used to predict bin centers without leveraging decoded features.

\section{Methodology}
\label{sec:methodology}
In this section, we describe our architecture, design choices and training protocol in detail.

\subsection{Overview}
We use the MiDaS~\cite{Ranftl2020MiDaS} training strategy for relative depth prediction. MiDaS uses a loss that is invariant to scale and shift. If multiple datasets are available, a multi-task loss that ensures pareto-optimality across the datasets is used. The MiDaS training strategy can be applied to many different network architectures. We use the DPT encoder-decoder architecture as our base model~\cite{Ranftl_2021_ICCV_DPT}, but replace the encoder with more recent transformer-based backbones \cite{DBLP:journals/corr/abs-2106-08254}. 
After pre-training the MiDaS model for relative depth prediction, we add one or more heads for metric depth estimation by attaching our proposed \emph{metric bins module} to the decoder (see Fig.~\ref{fig:architecture-zoe} for the overall architecture). The metric bins module outputs metric depth and follows the adaptive binning principle, originally introduced in~\cite{bhat2021adabins} and subsequently modified by~\cite{bhat2022localbins, li2022binsformer, agarwal2022attention, pixelbinsSarwari:EECS-2021-32}. In particular, we start out with the pixel-wise prediction design as in LocalBins~\cite{bhat2022localbins} and propose modifications that further improve performance. Finally, we fine-tune the complete architecture end-to-end. 

\subsection{Architecture Details}
We first review LocalBins, and then introduce our novel metric bins module with \emph{attractor layers}, our bin aggregation strategy, and loss function.
\paragraph{LocalBins review.} Our metric bins module is inspired by the LocalBins architecture proposed in~\cite{bhat2022localbins}. LocalBins uses a standard encoder-decoder as the base model and attaches a module that takes the multi-scale features from the encoder-decoder as input and predicts the bin centers at every pixel. Final depth at a pixel is obtained by a linear combination of the bin centers weighted by the corresponding predicted probabilities. The LocalBins module first predicts $N_{seed}$ different seed bins at each pixel position at the bottleneck. Each bin is then split into two at every decoder layer using splitter MLPs. The number of bin centers is doubled at every decoder layer and we end up with $2^nN_{seed}$ bins at each pixel at the end of $n$ decoder layers. Simultaneously, the probability scores (\textbf{p}) over $N_{total}=2^nN_{seed}$ bin centers (\textbf{c}) are predicted from the decoder features using softmax and the final depth at pixel $i$ is obtained using:
\begin{equation}
\label{eq:lin_comb}
    d(i) = \sum_{k=1}^{N_{total}}{p_i(k)c_i(k)}
\end{equation}

\begin{figure}[!htb]
    \includegraphics[width=\columnwidth]{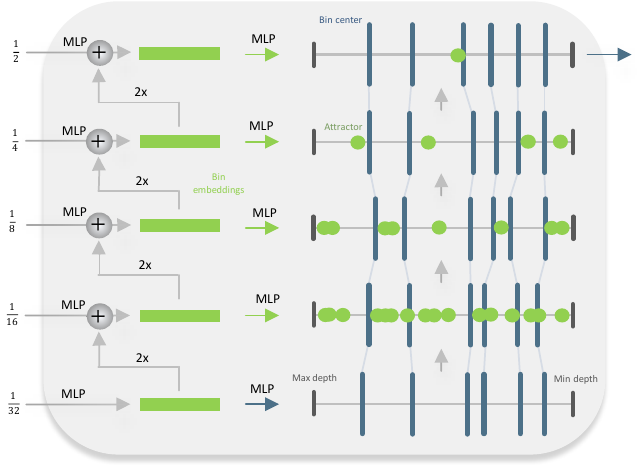}
    \vspace{-12pt}
    \caption{\textbf{Metric Bins Module.} Five incoming channels, corresponding to different depth hierarchies (see Fig.~\ref{fig:architecture-zoe}), are converted to 1-dimensional bin embeddings (green boxes) by MLPs, in combination with upsampling and addition operations. The lowest bin embedding yields metric bin centers (blue, vertical lines; not representative of actual number 64), whereas the remaining embeddings provide attractors for their respective hierarchy levels (green dots).
    Going upwards in the metric bins module, the attractors pull the bin centers according to Eqs.~(\ref{eq:attractor_fractional}) and (\ref{eq:attractor_exponential}). 
    }
    \label{fig:architecture-bin-module}
    \vspace{-12pt}
\end{figure}

\paragraph{Metric bins module.}
The metric bins module takes multi-scale features from the MiDaS decoder as input and predicts the bin centers to be used for metric depth prediction (see Fig.~\ref{fig:architecture-bin-module}). However, instead of starting with a small number of bins at the bottleneck and splitting them later, our metric bins module predicts all the bin centers at the bottleneck and adjusts them at subsequent decoder layers. This bin adjustment is implemented via our newly proposed building block, called \textit{attractor layers}.

\paragraph{Attract instead of split.}
LocalBins implements multi-scale refinement of the bins by splitting them conditioned on the multi-scale features. In contrast, we implement the multi-scale refinement of the bins by adjusting them, moving them left or right on the depth interval. Using the multi-scale features, we predict a set of points on the depth interval towards which the bin centers get attracted. More specifically, at the $l^{th}$ decoder layer, an MLP takes the features at a pixel as input and predicts $n_a$ attractor points $\{a_k: k = 1,...,n_a\}$ for that pixel position. The adjusted bin center is $c_i' = c_i + \Delta c_i$, with the adjustment given by: 
    \begin{align}
        \label{eq:attractor_fractional}
        \Delta c_i &= \sum_{k=1}^{n_a}\frac{a_k - c_i}{ 1 + \alpha |a_k - c_i|^\gamma}
    \end{align}
where the hyperparameters $\alpha$ and $\gamma$ determine the attractor strength. We name this attractor variant \textit{inverse attractor}. We also experiment with an exponential variant given by:
    \begin{align}
        \label{eq:attractor_exponential}
        \Delta c_i &= \sum_{k=1}^{n_a}(a_k - c_i)e^{-\alpha |a_k - c_i|^\gamma}
    \end{align}

Our experiments suggest that the \textit{inverse attractor} leads to better performance. 
We let the number of attractor points vary from one decoder layer to another, denoted together as a set $\{n^l_a\}$. We use $N_{total}=64$ bins and $\{16,8,4,1\}$ attractors. Please refer to Sec.~\ref{sec:ablation} for various ablations.

The attracting strategy is preferred because it's a contracting process while splitting is inherently dilative. Splitting adds extra constraints of newly produced bins summing up to the original bin width, while attractors adjust freely without such local constraints (only the total width is invariant). Intuitively, the prediction should get more refined and focused with decoder layers, which attractors achieve without dealing with any local constraints.

\paragraph{Log-binomial instead of softmax.}
To get the final metric depth prediction, the bin centers are linearly combined, weighted by their probability scores as per Eq.~(\ref{eq:lin_comb}). Prior adaptive bins based models~\cite{bhat2021adabins,bhat2022localbins,li2022binsformer,agarwal2022attention} use a softmax to predict the probability distribution over the bin centers. The choice of softmax is mainly inspired from the discrete classification analogy. Although the softmax plays well with unordered classes, since the bins are inherently ordered, it intuitively makes sense to use an ordering-aware prediction of the probabilities. The softmax approach can result in vastly different probabilities for nearby bin centers ({\scriptsize $|p_i - p_{i+1}|>>0$}).
Inspired by Beckham and Pal \cite{unimodal-pmlr-v70-beckham17a}, we use a binomial distribution instead to address this issue and correctly consider ordinal relationships between bins. 

The binomial distribution has one parameter $q$ which controls the placement of the mode. We concatenate the relative depth predictions with the decoder features and predict a 2-channel output ($q$ - mode and $t$ - temperature) from the decoder features to get the probability score over the $k^{th}$ bin center by: 
\begin{equation}
    p(k; N, q) = {N \choose k} q^k (1 - q)^{N-k}
\end{equation}
where $N=N_{total}$ is the total number of bins. In practice, since we use large values of $N$, we take $\log{(p)}$, use Stirling's approximation~\cite{abramowitz2002stegun} for factorials and apply $\text{softmax}(\{\log{(p_k)}/t\}_{k=1}^{N})$ to get normalized scores for numerical stability. The parameter $t$ controls the temperature of the resulting distribution. The softmax normalization preserves the unimodality of the logits. 
Finally, the resulting probability scores and the bin centers from the metric bins module are used to obtain the final depth as per Eq.~(\ref{eq:lin_comb}).

\paragraph{Loss.} We use the scale-invariant log loss ($\mathcal{L}_{pixel}$) for pixel-level supervision as in LocalBins \cite{bhat2022localbins}.
Unlike LocalBins, we do not use the chamfer loss for bins due to the high memory requirement but only limited improvement.

\subsection{Training strategies}
\label{subsec:train_strat}
As described previously, we have two stages for training: relative depth pre-training for the MiDaS backbone and metric depth fine-tuning for the prediction heads. We compare models with and without pre-training for relative depth as in~\cite{Ranftl2020MiDaS}. We also explore different variations of fine-tuning, using a single dataset and multiple datasets; in the case of multiple datasets, we also compare using a single head, \ie metric bins module, to using multiple heads. Please refer to \cref{sec:models} for more details about the exact model definitions. In the supplement, we report results for additional variations. 

\paragraph{Metric fine-tuning on multiple datasets}
Training a metric depth model on a mixture of datasets with a wide variety of scenes, for example from indoor and outdoor domains, is hard. The model not only has to handle images taken with different cameras and camera settings but also has to learn to adjust for the large variations in the overall scale of the scenes. Indoor scenes are usually limited to a maximum depth of 10 meters while outdoor scenes can have infinite depth (capped at 80 meters in most prior works). 
We hypothesize that a backbone pre-trained for relative depth estimation, alleviates the issues of fine-tuning on multiple datasets to some extent. 
We can also equip the model with multiple metric bins modules, one for each scene type (indoor versus outdoor). Different metric heads can be thought of as scene-type experts. Note that the base model is still common to all metric heads; the complete model with multiple heads is trained end-to-end. See \cref{sec:multi-dataset} for a comparison of our model with single head and multiple heads. 

\paragraph{Routing to metric heads.}
When the model has multiple metric heads, we need a router that chooses the metric head to use for a particular input. We employ commonly used routing mechanisms developed in other contexts, \eg, see Fedus \etal \cite{fedus2022review} for a review. We explore three main variants: {\textit{(R.1)}} \textit{Labeled Router:} In this variant, we provide scene type labels (indoor or outdoor) to the model at both training and inference times and manually map from the scene type to the metric head. {\textit{(R.2)}} \textit{Trained Router:} Here, we train a classifier MLP that predicts the scene type of the input image based on the bottleneck features and then routes to the corresponding metric head. Therefore, this variant only needs scene-type labels during training. {\textit{(R.3)}} \textit{Auto Router:} In this setting, a router MLP (equivalent to a classifier in {\textit{R.2}}) is used, but no labels are provided during either training or inference. Both the trainable router types, \textit{Trained Router} and \textit{Auto Router}, are trained end-to-end with the whole model. See \cref{sec:ablation} for a performance comparison of the discussed routing mechanisms.

\section{Experimental Setup}

\subsection{Datasets}
Our primary datasets for training ZoeDepth are NYU Depth v2 (N) for indoor and KITTI (K) for outdoor scenes. We refer to the combination of both datasets as (NK). For pre-training the relative depth backbone, we train on a mix of 12 datasets (M12) consisting of the 10 datasets used in \cite{Ranftl_2021_ICCV_DPT}: HRWSI~\cite{xian2020structure}, BlendedMVS~\cite{yao2020blendedmvs}, ReDWeb~\cite{xian2018monocular}, DIML-Indoor~\cite{kim2018deep}, 3D Movies~\cite{Ranftl2020MiDaS}, MegaDepth~\cite{MDLi18}, WSVD~\cite{wang2019web}, TartanAir~\cite{wang2020tartanair}, ApolloScape~\cite{huang2019apolloscape} and IRS~\cite{wang2019irs}, plus 2 additional datasets: KITTI~\cite{Menze_2015_CVPR} and NYU Depth v2~\cite{Silberman2012}.

To demonstrate generalizability, we evaluate zero-shot performance on a number of real-world and synthetic datasets: SUN RGB-D~\cite{Song2015_sunrgbd}, iBims~\cite{koch2019}, DIODE Indoor~\cite{diode_dataset} and HyperSim ~\cite{roberts:2021} for the indoor domain;
DDAD~\cite{packnet}, DIML Outdoor~\cite{kim2018deep}, DIODE Outdoor~\cite{diode_dataset} and Virtual KITTI~2~\cite{cabon2020vkitti2} for the outdoor domain.
We provide further details about the datasets in the supplement. 

\begin{figure*}[!ht]
    \centering
    \begin{overpic}[scale=.25,width=0.9\linewidth]{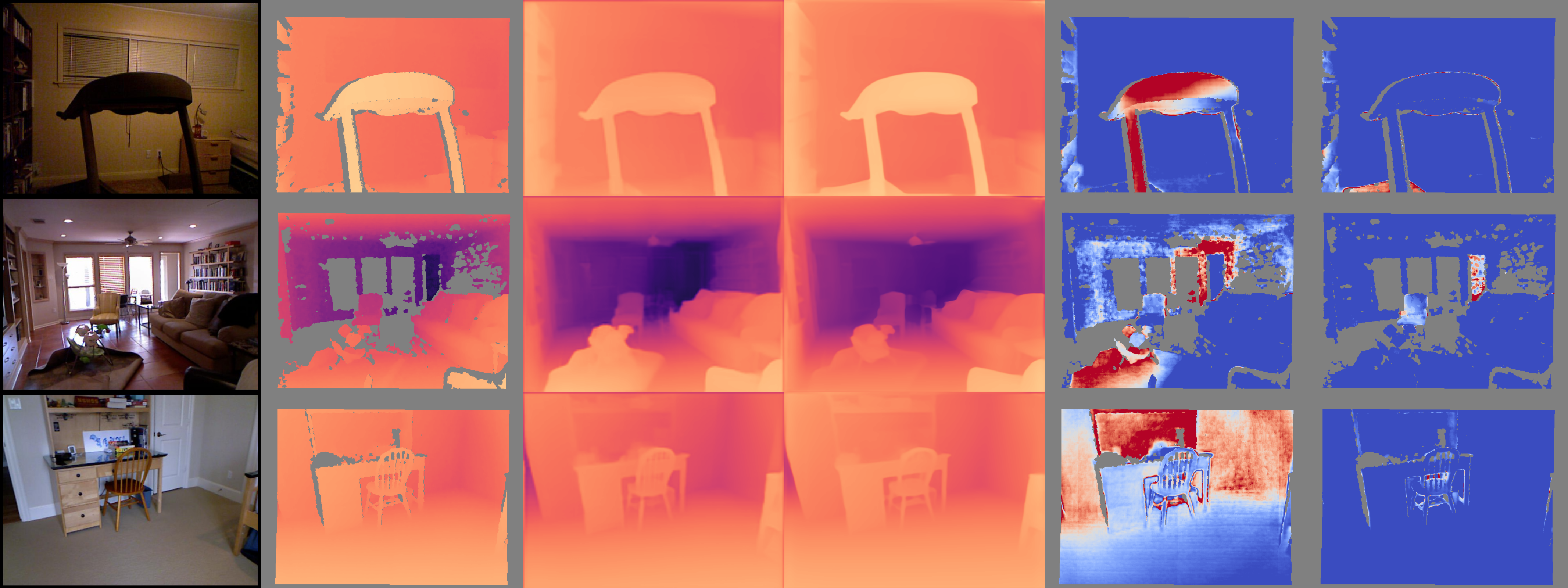}
    \put(6,-2){Input}
    \put(19,-2){Ground Truth}
    \put(35,-2){NeWCRFs~\cite{yuan2022new}}
    \put(55,-2){Ours}
    \put(70,-2){NeWCRFs $\Delta$}
    \put(87,-2){Ours $\Delta$}
    
    \end{overpic}
    \vspace{0.7em}
    \caption{\textbf{Qualitative comparison on NYU Depth v2.} Our method consistently produces better predictions with much less error. When looking closely at the depth maps it can also be observed that our predictions are much sharper with clear edges. $\Delta$ indicates square error ranging from lowest (dark blue) to highest (dark red) across predictions. Invalid regions are indicated as grey.}
    \label{fig:nyu_depth_qualitative}
    \vspace{-8pt}
\end{figure*}
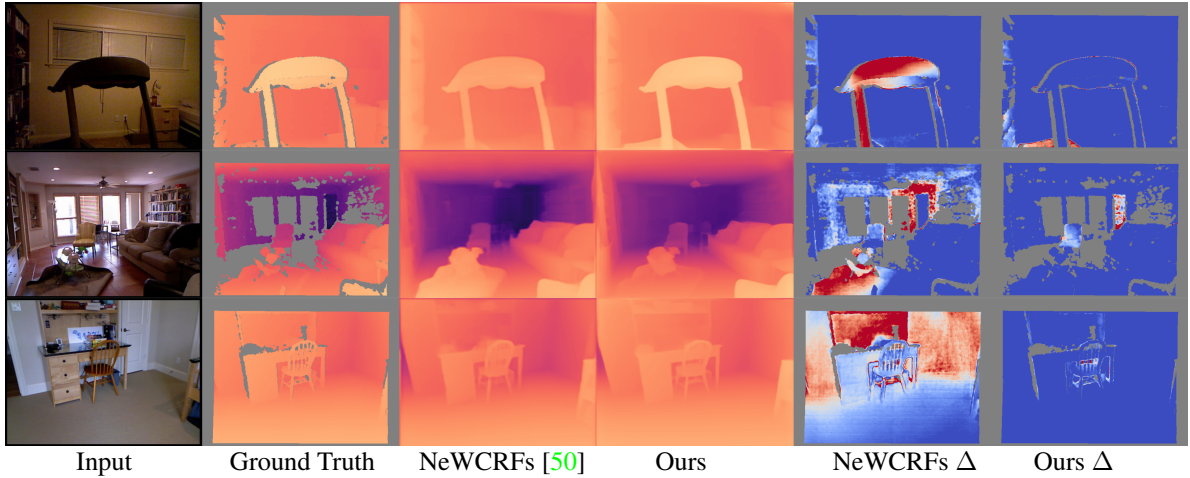

\subsection{Models}
\label{sec:models}
The models are named according to the following convention: \textit{ZoeD-\{RDPT\}-\{MFT\}}, where \textit{ZoeD} is the abbreviation for ZoeDepth, \textit{RDPT} denotes the datasets used for relative depth pre-training (X denotes no pre-training) and \textit{MFT} denotes the datasets used for metric depth fine-tuning. We train and evaluate the following models: \textit{ZoeD-X-N}, \textit{ZoeD-X-K}, \textit{ZoeD-M12-N},\textit{ ZoeD-M12-K} and\textit{ ZoeD-M12-NK}. All models use the BEiT\textsubscript{384}-L backbone from timm \cite{rw2019timm} that was pre-trained on ImageNet. The models \textit{ZoeD-X-N} and \textit{ZoeD-X-K} are directly fine-tuned for metric depth on NYU Depth v2 and KITTI respectively without any pre-training for relative depth estimation.\textit{ ZoeD-M12-N} and \textit{ZoeD-M12-K} additionally include pre-training for relative depth estimation on the M12 dataset mix before the fine-tuning stage for metric depth. \textit{ZoeD-M12-NK} is also pre-trained on M12, but has two separate heads fine-tuned on both NYU Depth v2 and KITTI.\textit{ ZoeD-M12-NK$^\dagger$} is a variant of this model with a single head but otherwise the same pre-training and fine-tuning procedure. In the supplement, we provide further results for models trained on additional dataset combinations in pre-training and fine-tuning.

\subsection{Evaluation Metrics}

We evaluate in metric depth space $\mathbf{d}$ by computing the absolute relative error (REL) $= \frac{1}{M}\sum_{i=1}^{\put(0,-3){\scriptsize{$M$}}}|\mathbf{d}_{i}-\hat{\mathbf{d}}_{i}|/\mathbf{d}_{i}$, the root mean squared error (RMSE) $= [\frac{1}{M}\sum_{i=1}^{\put(0,-3){\scriptsize{$M$}}}|\mathbf{d}_{i}-\hat{\mathbf{d}}_{i}|^{2}]^\frac{1}{2}$, the average $\log_{10}$ error $= \frac{1}{M}\sum_{i=1}^{\put(0,-3){\scriptsize{$M$}}}|\log_{10}{\mathbf{d}_{i}}-\log_{10}{\hat{\mathbf{d}}_{i}}|$, and the threshold accuracy $\delta_n = \%$ of pixels s.t.~$\max{(\mathbf{d}_{i}/\hat{\mathbf{d}}_{i},\hat{\mathbf{d}}_{i}/\mathbf{d}_{i})} < 1.25^n$ for $n=1,2,3$, where $\mathbf{d}_{i}$ and $\hat{\mathbf{d}}_{i}$ refer to ground truth and predicted depth at pixel $i$, respectively, and $M$ is the total number of pixels in the image. We cap the evaluation depth at 10m for indoor datasets (8m for SUN RGB-D) and at 80m for outdoor datasets. 
Final model outputs are computed as the average of an image's depth prediction and the prediction of its mirror image and are evaluated at ground truth resolution.

In addition, we define two metrics to measure relative improvement ($\text{RI}$) across datasets and metrics respectively. For $M$ datasets $D_i$ with $i \in{[1,M]}$, we compute $\text{mRI}_{D} = \frac{1}{M}\sum_{i=1}^M \text{RI}_{D_i} $. Similarly, for $N$ metrics $\theta_j$ with $j \in{[1,N]}$, we compute $\text{mRI}_{\theta} = \frac{1}{N}\sum_{j=1}^N \text{RI}_{\theta_i} $. For metrics where lower is better, $\text{RI} = \frac{r-t}{r}$ and for metrics where higher is better $\text{RI} = \frac{t-r}{r}$, where $r$ and $t$ correspond to the reference and target scores respectively.

\section{Results}
\label{sec:results}

\subsection{Comparison to SOTA on NYU Depth V2}

\begin{table}[!b]
\centering
\footnotesize
\begin{tabularx}{\linewidth}{@{}l|*{5}{C}c@{}}
\toprule
\textbf{Method} &
\textbf{$\delta_1$}$\uparrow$       & \textbf{$\delta_2$}$\uparrow$          & \textbf{$\delta_3$}~$\uparrow$            & REL~$\downarrow$          & RMSE~$\downarrow$  & $\log_{10}$~$\downarrow$ \\ \midrule
Eigen~\etal~\cite{Eigen2014}    & 0.769          & 0.950          & 0.988          & 0.158            & 0.641          & --              \\  
Laina~\etal~\cite{Laina2016}& 0.811          & 0.953          & 0.988          & 0.127            & 0.573          & 0.055                \\
Hao~\etal~\cite{Hao2018DetailPD}        & 0.841          & 0.966          & 0.991          & 0.127            & 0.555          & 0.053                \\ 
DORN~\cite{Fu2018DeepOR}   & 0.828          & 0.965          & 0.992          & 0.115            & 0.509          &   0.051  \\
SharpNet~\cite{Ramamonjisoa_2019_ICCV}          & 0.836          & 0.966          & 0.993          & 0.139            & 0.502          &      {0.047}          \\
Hu~\etal~\cite{Hu2018RevisitingSI}        & 0.866          & 0.975          & 0.993          & 0.115            & 0.530          &    0.050            \\ 
Lee~\etal~\cite{Lee2011}    & 0.837          & 0.971          & 0.994          & 0.131            & 0.538          &   --  \\
Chen~\etal~\cite{ijcai2019-98}        & 0.878          & 0.977          & 0.994          & 0.111            & 0.514          &  0.048              \\ 
BTS~\cite{bts_lee2019big}& {0.885}          & 0.978          & 0.994          & 0.110            & {0.392}          & {0.047}          \\ 
Yin~\etal~\cite{Yin_2019_ICCV}   & 0.875          & 0.976          & 0.994          & {0.108}            & 0.416          & 0.048               \\ 
AdaBins~\cite{bhat2021adabins} & {0.903} & {0.984} & {0.997} & {0.103}     & {0.364} & {0.044} \\ 
{LocalBins~\cite{bhat2022localbins}}  & {0.907} & {0.987} & \underline{0.998} & {0.099}     & {0.357} & {0.042} \\ 
Jun~\etal~\cite{jun2022depth}  & 0.913 & 0.987 & \underline{0.998} & {0.098}     & {0.355} & {0.042} \\ 
{NeWCRFs~\cite{yuan2022new}}  & {0.922} & 0.992 & \underline{0.998} & {0.095}     & {0.334} & {0.041} \\ 
\midrule
\textbf{ZoeD-X-N}   & 0.946 & \underline{0.994} & \textbf{0.999} & {0.082}     & 0.294 & 0.035 \\
\textbf{ZoeD-M12-N}  & \textbf{0.955} & \textbf{0.995} & \textbf{0.999} & \textbf{0.075}     & \textbf{0.270} & \textbf{0.032} \\ 
\midrule
\textbf{ZoeD-M12-NK}  & \underline{0.953} & \textbf{0.995} & \textbf{0.999} & \underline{0.077}     & \underline{0.277} & \underline{0.033} \\ 
\bottomrule
\end{tabularx}
\vspace{-6pt}
\caption{\textbf{Quantitative comparison on NYU-Depth v2.} The reported numbers of prior art are from the corresponding original papers. Best results are in bold, second best are underlined. }
\label{tab:results-nyu}
\end{table}

Our novel architecture beats SOTA without using any additional data for pre-training.
To demonstrate this, we evaluate our model \textit{ZoeD-X-N} on the popular metric depth estimation benchmark NYU Depth v2~\cite{Silberman2012}.
\textit{ZoeD-X-N} is not pre-trained for relative depth; the backbone is initialized with the standard weights from ImageNet pre-training.
\cref{tab:results-nyu} shows the performance of models on the official NYU Depth v2 test set.  This model already outperforms NeWCRFs~\cite{yuan2022new} by 13.7\% (REL = 0.082), highlighting the contribution of our architecture design. 

Next, we verify that our architecture can benefit from relative depth pre-training. Our corresponding model \textit{ZoeD-M12-N} significantly outperforms the prior state-of-the-art NeWCRFs~\cite{yuan2022new} by nearly 21\% (REL = 0.075). Results are not just numerically better; the resulting depth maps also have much sharper boundaries (see \cref{fig:nyu_depth_qualitative}). 
We believe this is the first demonstration of successful relative depth pre-training at a competitive level.
While other architectures can also benefit from pre-training, some modifications are required.
In the next section, we show one such modification by combining our base model with architecture building blocks proposed by other papers (see \cref{tab:results-nk}). This shows that while other architectures benefit from our larger backbone and relative depth pre-training, they are still not competitive with our complete framework.

\subsection{Universal Metric SIDE}
\label{sec:multi-dataset}
Here, we evaluate our progress towards a universal metric depth estimation framework by analyzing our model \textit{ZoeD-M12-NK} which was trained across two different metric datasets and generalizes across indoor and outdoor domains.
Models trained across multiple metric datasets usually perform worse or diverge. In contrast, our model \textit{ZoeD-M12-NK} still outperforms the previous SOTA NeWCRFs~\cite{yuan2022new} on NYU Depth v2 by 18.9\% (REL = 0.077, \cref{tab:results-nyu}).
 While \textit{ZoeD-M12-NK} is not as good as our model (\textit{ZoeD-M12-N}) fine-tuned only on NYU Depth v2, it provides a very attractive trade-off between performance and generalization across domains.
 
 To underline the difficulty of cross-domain training, we perform a comparison to other models trained simultaneously on indoor and outdoor datasets (NK). First, we evaluate recent methods trained on NK without any architectural modifications and compare them with our method in \cref{tab:results-nk}. 
We find that existing works are unable to achieve competitive results in that setting. AdaBins~\cite{bhat2021adabins} and PixelBins~\cite{pixelbinsSarwari:EECS-2021-32} fail to converge at all, while the SOTA NeWCRFs'~\cite{yuan2022new} performance degrades by nearly 15\% (REL 0.095 to 0.109) on NYU (compare \cref{tab:results-nk} with \cref{tab:results-nyu}). These experiments confirm that previous models significantly degrade when being trained jointly on datasets from different domains.
In contrast, we only observe an 8\% drop (REL 0.075 to 0.081) while using the shared head, demonstrating our model's ability to deal with different domains at once. This gap is further reduced to mere 2.6\% (REL 0.075 vs 0.077) using our two-head model \zoedmnk{}, outperforming NeWCRFs~\cite{yuan2022new} by 25.2\% (mRI$_D$ in REL).  

\begin{table}[bt]
    \centering
    \setlength{\tabcolsep}{4pt} 
    \footnotesize
    \begin{tabular}{@{}lcccc|c@{}}
    \toprule
    \textbf{Method}           & \textbf{NYU}              & \textbf{KITTI}      & \textbf{iBims-1}                     & \textbf{vKITTI-2} & \textbf{mRI$_D$} \\ 
    \midrule
    \multicolumn{6}{l}{\textbf{Baselines: no modification}} \\\midrule
    \nomod{DORN}        &    0.156             &    0.115       &           0.287                &    0.259           &   -45.7\%          \\
    \nomod{LocalBins}   &   0.245   &   0.133       &        0.296                   &         0.265               &  -74.0\%  \\
    \nomod{PixelBins}   &      -           &    -    &         -                   &         -           &    -    \\
    \nomod{NeWCRFs}     &  0.109   &    0.076       &     {0.189}                      &    0.190     & 0.0\%                 \\
    \midrule
    \multicolumn{6}{l}{\textbf{Baselines: modified to use our pre-trained DPT-BEiT-L as backbone}} \\\midrule
    \modified{DORN}        & 0.110                 & 0.081          & 0.242                           & 0.215               &    -12.2\%         \\
    \modified{LocalBins}   & {0.086}     & {0.071}          & {0.221}                           & 0.121                        & 11.8\%   \\
    \modified{PixelBins}   & 0.088                 & {0.071}          & 0.232                           & {0.119}                    &   10.1\%     \\
    \modified{NeWCRFs}     & 0.088          & 0.073          & 0.233                           & 0.124           &  8.7\%              \\
    \midrule
    \multicolumn{6}{l}{\textbf{Ours: different configuations for fair comparison}} \\\midrule
    {$\text{ZoeD-X-NK}^\dagger$}        &   0.095      &    0.074      &          \underline{0.187}                 & 0.184 & 4.9\%\\
    {$\text{ZoeD-M12-NK}^\dagger$}        & \underline{0.081}        & \underline{0.061}         & {0.210}                          & \underline{0.112} & \underline{18.8\%} \\   
    \textbf{$\text{ZoeD-M12-NK}$}        & \textbf{0.077}        & \textbf{0.057}         & \textbf{0.186}                          & \textbf{0.105} & \textbf{25.2\%} \\ 
    \bottomrule
    \end{tabular}
    \caption{\textbf{Comparison with existing works when trained on NYU and KITTI.} Results are reported using the REL metric. The mRI$_D$ column denotes the mean relative improvement with respect to NeWCRFs across datasets.
    X in the model name, means no architecture change and no pre-training. M12 means that the model was pre-trained (using our base model based on the DPT architecture with the BEiT-L encoder).
    All models are fine-tuned on NYU and KITTI. 
    $\dagger$~denotes a single metric head (shared); single-head training allows us to adapt prior models without major changes. Best results are in bold, second best are underlined.
    PixelBins~\cite{pixelbinsSarwari:EECS-2021-32} did not converge without modification. We also tried to train AdaBins \cite{bhat2021adabins} across both datasets, but despite our best effort and extensive hyperparameter tuning, it did not converge.}
    \label{tab:results-nk}
\vspace{-6mm}
\end{table}

We conclude that previous models require changes to successfully train on multiple datasets. We conduct additional experiments where we improve previous models by incorporating part of our framework. Specifically, we use the same base model (DPT with a BEiT-L backbone), with relative pre-training on M12, and with fine-tuning on NK (NYU and KITTI mixture) using a single metric head; only the design of the metric head varies.
\cref{tab:results-nk} shows that previous models still fall behind in this setting.
Since DORN~\cite{Fu2018DeepOR} and LocalBins~\cite{bhat2022localbins} are light-weight and modular, they can be easily used in conjuction with our pre-trained relative depth model instead of our metric bins module. NeWCRFs~\cite{yuan2022new} is originally a tightly-coupled decoder-focused design; however, to keep the base model exactly the same, we use an extra head with NeWCRFs layers that use DPT decoder features as multi-scale input. This increases the complexity significantly ($\sim$40M more parameters than ours) yet still underperforms when compared to pixel-wise bins-based methods: LocalBins, PixelBins and \textit{ZoeD-M12-NK}. This suggests that bins-based architectures are better suited for multi-domain training and
can better exploit relative depth pre-training. Our model performs best both on NYU and KITTI, as well as on iBims-1 and virtual KITTI-2 that have not been seen in training. These results indicate that our metric bins module exploits pre-training better than existing works, enabling improved domain adaptation and generalization (zero-shot performance). We investigate zero-shot performance in more detail next.

\begin{table*}[!htb]
\footnotesize
\centering
\setlength{\tabcolsep}{3pt} 
\begin{tabular}
{@{}
    l@{\hspace{6pt}}
    c@{\hspace{6pt}}c@{\hspace{3pt}}c@{\hspace{3pt}}|c@{\hspace{3pt}}|
    c@{\hspace{6pt}}c@{\hspace{3pt}}c@{\hspace{3pt}}|c@{\hspace{3pt}}|
    c@{\hspace{6pt}}c@{\hspace{3pt}}c@{\hspace{3pt}}|c@{\hspace{3pt}}|
    c@{\hspace{6pt}}c@{\hspace{3pt}}c@{\hspace{3pt}}|c@{\hspace{3pt}}
    @{}}
\toprule
& \multicolumn{4}{c|}{\textbf{SUN RGB-D}} & \multicolumn{4}{c|}{\textbf{iBims-1 Benchmark}} & \multicolumn{4}{c|}{\textbf{DIODE Indoor}} & \multicolumn{4}{c}{\textbf{HyperSim}}\\

Method & $\delta_{1}$\,$\uparrow$ & REL\,$\downarrow$ & RMSE\,$\downarrow$ & mRI$_\theta$\,$\uparrow$ & $\delta_{1}$\,$\uparrow$ & REL\,$\downarrow$ & RMSE\,$\downarrow$ & mRI$_\theta$\,$\uparrow$ & $\delta_{1}$\,$\uparrow$ & REL\,$\downarrow$ & RMSE\,$\downarrow$ & mRI$_\theta$\,$\uparrow$ & $\delta_{1}$\,$\uparrow$ & REL\,$\downarrow$ & RMSE\,$\downarrow$ & mRI$_\theta$\,$\uparrow$   \\ 
\midrule
BTS~\cite{bts_lee2019big} & 0.740 & 0.172& 0.515 & -14.2\% & 0.538 & 0.231& 0.919 & -6.9\%  & 0.210 & 0.418& 1.905 & 2.3\%& 0.225 & 0.476& 6.404 & -8.6\% \\
AdaBins~\cite{bhat2021adabins}  & 0.771 & 0.159& 0.476 & -7.0\%  & 0.555 & 0.212& 0.901 & -2.1\%  & 0.174 & 0.443& 1.963 & -7.2\%  & 0.221 & 0.483& 6.546 & -10.5\%\\
LocalBins~\cite{bhat2022localbins} & 0.777 & 0.156& 0.470 & -5.6\%  & 0.558  & 0.211 & 0.880  & -0.7\%  & 0.229  & 0.412 & 1.853 & 7.1\%
  & 0.234  & 0.468 & 6.362 & -6.6\%  \\
NeWCRFs~\cite{yuan2022new} & 0.798 & 0.151& 0.424 & 0.0\%& 0.548 & 0.206& 0.861 & 0.0\%& 0.187 & 0.404& 1.867 & 0.0\%& 0.255 & 0.442& 6.017 & 0.0\%  \\

\midrule
\textbf{ZoeD-X-N} & \underline{0.857}  & {0.124} & 0.363 & {13.2\%}& \textbf{0.668}  & \underline{0.173} & \underline{0.730} & \underline{17.7\%}& \textbf{0.400}  & \textbf{0.324} & \textbf{1.581} & \textbf{49.7\%}& \underline{0.284}  & 0.421 & 5.889 & \underline{6.1\%}  \\

\textbf{ZoeD-M12-N}  & \textbf{0.864} & \textbf{0.119} & \textbf{0.346} &     \textbf{16.0\%} & \underline{0.658} &   \textbf{0.169} &    \textbf{0.711} &     \textbf{18.5\%} & {0.376} &  \underline{0.327} &   \underline{1.588} &  {45.0\%} & \textbf{0.292} &  \textbf{0.410} &   \textbf{ 5.771} &      \textbf{8.6\%}\\

\midrule
\textbf{ZoeD-M12-NK} & 0.856 &  \underline{0.123} & \underline{0.356} & \underline{13.9\%} & 
0.615 & 0.186 & 0.777 & 10.6\% & \underline{0.386} &  0.331 &   1.598 &       \underline{46.3\%} &   0.274 &    \underline{0.419} & \underline{5.830} &  5.3\% \\
\bottomrule
\end{tabular}
\vspace{-6pt}
\caption{\textbf{Quantitative results for zero-shot transfer to four unseen indoor datasets.} mRI$_\theta$ denotes the mean relative improvement with respect to NeWCRFs across all metrics ($\delta_1$, REL, RMSE). Evaluation depth is capped at 8m for SUN RGB-D, 10m for iBims and DIODE Indoor, and 80m for HyperSim. Best results are in bold, second best are underlined.}
\label{tab:zero-shot-indoors}
\end{table*}

\begin{table*}[!htb]
\footnotesize
\centering
\setlength{\tabcolsep}{2.5pt} 
\begin{tabular}{@{}
    l@{\hspace{6pt}}
    c@{\hspace{6pt}}c@{\hspace{2.5pt}}c@{\hspace{2.5pt}}|c@{\hspace{2.5pt}}|
    c@{\hspace{6pt}}c@{\hspace{2.5pt}}c@{\hspace{2.5pt}}|c@{\hspace{2.5pt}}|
    c@{\hspace{6pt}}c@{\hspace{2.5pt}}c@{\hspace{2.5pt}}|c@{\hspace{2.5pt}}|
    c@{\hspace{6pt}}c@{\hspace{2.5pt}}c@{\hspace{2.5pt}}|c@{\hspace{2.5pt}}
    @{}}
\toprule
& \multicolumn{4}{c|}{\textbf{Virtual KITTI 2}} & \multicolumn{4}{c|}{\textbf{DDAD}} & \multicolumn{4}{c|}{\textbf{DIML Outdoor}} & \multicolumn{4}{c}{\textbf{DIODE Outdoor}} \\
Method & $\delta_{1}$\,$\uparrow$ & REL\,$\downarrow$ & RMSE $\downarrow$ & mRI$_\theta$\,$\uparrow$ & $\delta_{1}$\,$\uparrow$ & REL\,$\downarrow$ & RMSE $\downarrow$ & mRI$_\theta$\,$\uparrow$ & $\delta_{1}$\,$\uparrow$ & REL\,$\downarrow$ & RMSE $\downarrow$ & mRI$_\theta$\,$\uparrow$ & $\delta_{1}$\,$\uparrow$ & REL\,$\downarrow$ & RMSE $\downarrow$ & mRI$_\theta$\,$\uparrow$\\ 
\midrule
BTS~\cite{bts_lee2019big} & 0.831 & 0.115 & 5.368 & 2.5\% & 0.805 & 0.147 & 7.550 & -17.8\% & \underline{0.016} & 1.785 & \underline{5.908} & \underline{24.3\%} & 0.171 & 0.837 & 10.48 & -4.8\% \\
AdaBins~\cite{bhat2021adabins} & 0.826 & 0.122 & 5.420 & 0.0\% & 0.766 & 0.154 & 8.560 & -26.7\% & 0.013 & 1.941 & 6.272 & 9.7\% & 0.161 & 0.863 & 10.35 & -7.2\% \\

LocalBins~\cite{bhat2022localbins} & 0.810 & 0.127 & 5.981 & -5.3\% & 0.777 & 0.151 & 8.139 & -23.2\% & \underline{0.016} & 1.820 & 6.706 & 19.5\% & 0.170  &  {0.821} & 10.27 & -3.6\% \\
NeWCRFs~\cite{yuan2022new} &  0.829 & 0.117 & 5.691 & 0.0\% & \textbf{0.874} & \textbf{0.119} & \textbf{6.183} & \textbf{0.0\%} & 0.010 & 1.918 & 6.283 & 0.0\% & 0.176 & 0.854 & 9.228 & 0.0\%\\
\midrule
\textbf{ZoeD-X-K} & 0.837 & 0.112 & {5.338} & 3.8\% & 0.790 & 0.137 & 7.734 & -16.6\% & 0.005 & \underline{1.756} & 6.180 & -13.3\% & \underline{0.242} &  \underline{0.799} & {7.806} & \underline{19.8\%} \\
\textbf{ZoeD-M12-K}   & \textbf{0.864} &  \textbf{0.100}  & \textbf{4.974} & \textbf{10.5\%} &  \underline{0.835} & \underline{0.129} & \underline{7.108} & \underline{-9.3\%} & 0.003 & 1.921 &  6.978 & -27.1\% & \textbf{0.269} &     0.852 & \textbf{6.898} & \textbf{26.1\%}\\
\midrule
\textbf{ZoeD-M12-NK}  &  \underline{0.850} & \underline{0.105} &    \underline{5.095} &  \underline{7.8\%} &  0.824 & 0.138 & 7.225 &-12.8\% & \textbf{0.292} & \textbf{ 0.641} &  \textbf{3.610} &  \textbf{976.4\% } &  0.208 & \textbf{0.757} & \underline{7.569} & 15.8\% \\
\bottomrule
\end{tabular}
\vspace{-6pt}
\caption{\textbf{Quantitative results for zero-shot transfer to four unseen outdoor datasets.} mRI$_\theta$ denotes the mean relative improvement with respect to NeWCRFs across all metrics ($\delta_1$, REL, RMSE). Best results are in bold, second best are underlined.}
\label{tab:zero-shot-outdoors}
\end{table*}

\subsection{Zero-shot Generalization}
We evaluate the generalization capabilities of our approach by comparing its zero-shot performance to prior works on eight unseen indoor and outdoor datasets without fine-tuning; we show qualitative results in \cref{fig:intro} and report quantitative results in \cref{tab:zero-shot-indoors} and \cref{tab:zero-shot-outdoors}. 

\cref{tab:zero-shot-indoors} reports zero-shot generalization on indoor datasets. Even with fine-tuning across both the indoor (NYU Depth v2) and outdoor (KITTI) domains, our model \textit{ZoeD-M12-NK} demonstrates significantly better performance than previous state-of-the-art models. The mean relative improvement (mRI$_\theta$) ranges from 5.3\% for HyperSim to 46.3\% for DIODE Indoor. As expected, fine-tuning only on NYU Depth v2 so that the training and test domains are both indoor, \ie \textit{ZoeD-M12-N}, leads to an increase in mRI$_\theta$ on all datasets.
\textit{ZoeD-X-N} scores lower in most datasets due to the lack of relative depth pre-training.

\cref{tab:zero-shot-outdoors} reports zero-shot generalization on outdoor datasets. Similar as for the indoor datasets, pre-training on M12 is generally beneficial. \textit{ZoeD-M12-NK} improves from 7.8\% for Virtual KITTI 2 to 976.4\% for DIML Outdoor over NeWCRFs \cite{yuan2022new}. On DDAD, \textit{ZoeD-M12-NK} performs 12.8\% worse while NeWCRFs \cite{yuan2022new} is best.
The metrics in \cref{tab:zero-shot-outdoors} and the rightmost image in \cref{fig:intro} show the quality of our results. 
Overall, our framework is the top performer in 7 out of 8 datasets.

Probably the most interesting result is the high $\text{mRI}_{\theta}$ value of 976.4\% that \textit{ZoeD-M12-NK} achieves on DIML Outdoor. All other models are fine-tuned only on KITTI with large depth ranges but the DIML Outdoor dataset contains mainly close-up images of outdoor scenarios making it more similar to an indoor dataset. Since \textit{ZoeD-M12-NK} was also fine-tuned on NYU Depth v2 and automatically routes inputs to different heads, it seems to leverage its knowledge of the indoor domain to improve predictions. This is also supported by the low performance of \textit{ZoeD-X-K} and \textit{ZoeD-M12-K} which were only fine-tuned on KITTI. This result clearly shows the benefit of models fine-tuned across multiple domains for generalization to arbitrary datasets. We expect that defining more granular domains and fine-tuning a variant of our model with more than two heads across many metric datasets would lead to even better generalization performance.

\subsection{Ablation Studies}
\label{sec:ablation}
In this section we study the importance of various design choices in our models.

\begin{figure}[!htb]
    \centering
    \includegraphics[width=\columnwidth]{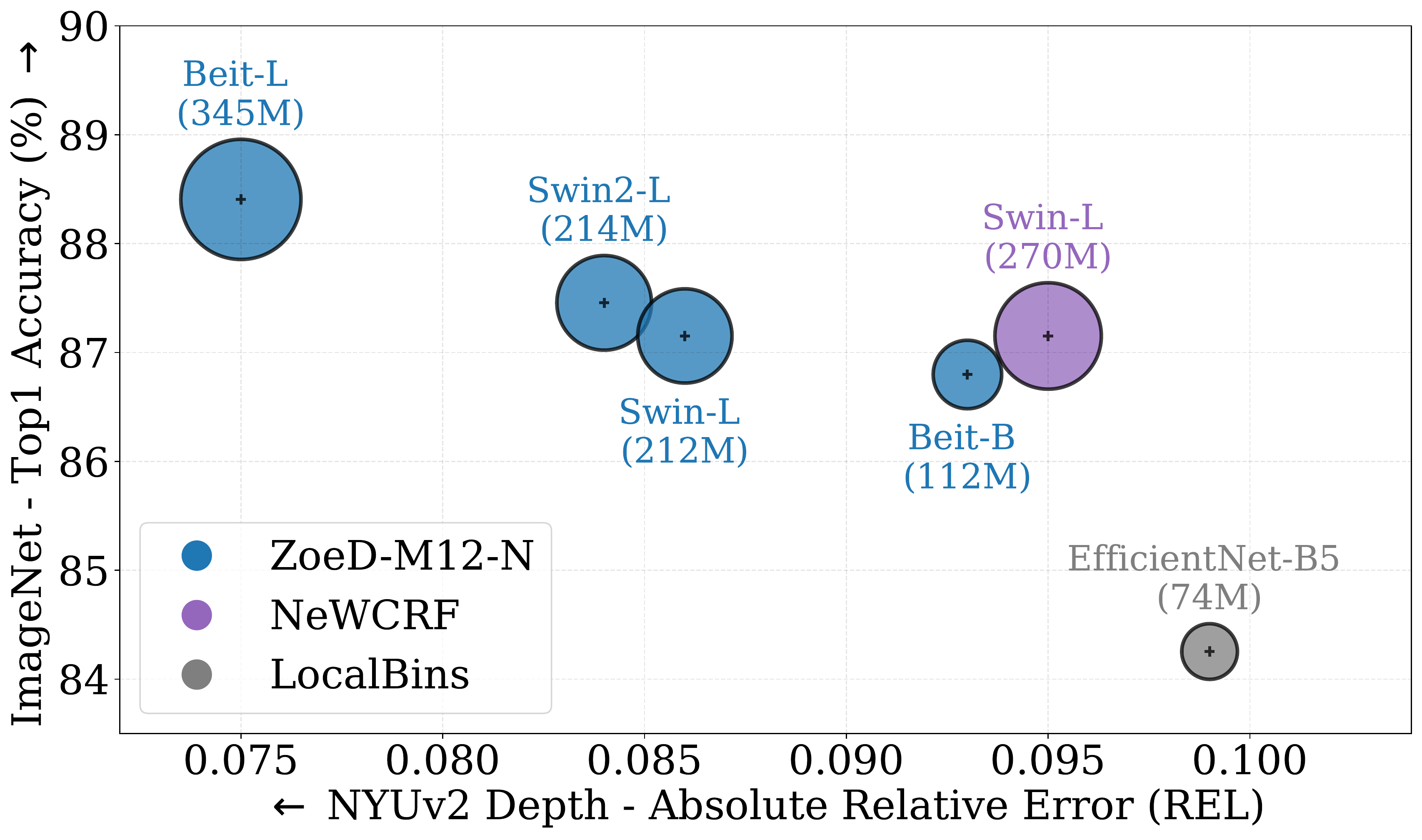}
    \vspace{-12pt}
    \caption{\textbf{Backbone ablation study.} There is a strong correlation between backbone performance in image classification and depth estimation. Larger backbones achieve lower absolute relative error (REL); with the same backbone and overall fewer parameters, our method still outperforms the current state-of-the-art NeWCRFs. The area of the circles is proportional to the number of parameters. The backbones shown are BEiT~\cite{DBLP:journals/corr/abs-2106-08254}, Swin2~\cite{liu2022swin}, Swin~\cite{liu2021swin} and EfficientNet B5~\cite{efficentnet_TanL19}, where L stands for large and B for base.}
    \label{fig:ablation-backbone}
    \vspace{-12pt}
\end{figure}

\paragraph{Backbones.} We study the effect of using different backbones for our base MiDaS model. The results are summarized in \cref{fig:ablation-backbone}. We find that larger backbones with more parameters lead to improved performance, but our model still outperforms the previous state of the art when using the same backbone \cite{liu2021swin}. Further, the image classification performance of the backbone is highly correlated to the performance of our depth estimation model, \ie lower absolute relative error (REL).  
Hence, our architecture can directly benefit from new backbones as they get introduced in the future.

\begin{table}[!htb]
\footnotesize
\setlength{\tabcolsep}{6.5pt} 
\begin{tabular}{ccc|cc}
\toprule
\multicolumn{3}{c|}{\textbf{Metric head type}}                                                              & \textbf{REL $\downarrow$} & \textbf{RMSE $\downarrow$} \\ \midrule
\multicolumn{1}{c|}{\textbf{Type}}                & \multicolumn{1}{c|}{\textbf{Variant}} & \textbf{Config} &              &              \\ \hline
\multicolumn{1}{c|}{Naive head}                   & \multicolumn{1}{c|}{-}         & \textbf{-}      & 0.096        & 0.335        \\ \hline
\multicolumn{1}{c|}{\multirow{6}{*}{Metric bins}} & \multicolumn{1}{c|}{Splitter}         & factor = 2      & 0.085        & 0.301        \\ 
\cline{2-5} 
\multicolumn{1}{c|}{} &
  \multicolumn{1}{c|}{\begin{tabular}[c]{@{}c@{}}Exponential\\ Attractor\end{tabular}} &
  \{16,8,4,1\} &
  0.086 &
  0.305 \\
  \cline{2-5} 
\multicolumn{1}{c|}{} &
  \multicolumn{1}{c|}{\multirow{4}{*}{\begin{tabular}[c]{@{}c@{}}Inverse\\ Attractor\end{tabular}}} &
  \{8,8,8,8\}     & 0.081        & 0.295        \\
\multicolumn{1}{c|}{}                             & \multicolumn{1}{c|}{}                 & \{16,2,2,16\}   & 0.081        & 0.291        \\ 
\multicolumn{1}{c|}{}                             & \multicolumn{1}{c|}{}                 & \{1,4,8,16\}    & \underline{0.080}        & \underline{0.287}        \\
\multicolumn{1}{c|}{}
& \multicolumn{1}{c|}{} &
  \{16,8,4,1\} &
  \textbf{0.075} &
  \textbf{0.270} \\
  \bottomrule
\end{tabular}
\vspace{-6pt}
\caption{\textbf{Metric head variants.} The ``\textbf{Config}" column specifies the split factor in case of the splitter variant and the number of attractors $\{n_a^l\}$ for attractor variants. The reported results are all based on \textit{ZoeD-M12-N} evaluated on NYU Depth v2. Best results are in bold, second best are underlined.}
\label{tab:ablation-metric-bins}
\vspace{-8pt}
\end{table}
\paragraph{Metric Bins Module.}
We study the contribution to the overall performance by various variants of our metric bins module listed in \cref{tab:ablation-metric-bins}.
First, we remove our metric bins module and attach a convolutional block to the decoder features from the base DPT model and directly predict the metric depth (standard regression). We call this variant \textit{naive head}. Our best attractor variant performs about 21\% better than the naive head. Notably, the metric bins with the splitter design as in \cite{bhat2022localbins} improves upon the naive head by 11.4\%, which is consistent with the 10.8\% improvement observed by \cite{bhat2022localbins} when comparing a naive Unet design with the splitter LocalBins design (refer to Tab.~3 in \cite{bhat2022localbins}). 
Next, we compare our novel attractor design with the splitter design of LocalBins~\cite{bhat2022localbins}. Our best attractor variant performs 11.7\% better. All the \textit{inverse} attractor variants perform decisively better than the splitter variant while the \textit{exponential} variant performs slightly worse. 

\paragraph{Routers.}
As discussed in \cref{subsec:train_strat}, we test the three variants for routing the relative features to metric heads. The results for the models with two metric heads, one for NYU Depth v2 and one for KITTI, are provided in \cref{tab:ablation-router}. Out of the three variants, the Auto Router performs the worst. This is expected as in this case the router never sees any domain labels. Surprisingly, the Trained Router performs
better on NYU Depth v2 than the Labeled Router, even though domain labels are unavailable during inference. We hypothesize that the domain-level discriminatory supervision may help in learning better representations. As we aim for a generic model without special requirements during inference, we choose the Trained Router and use it in all our multi-head models.

\paragraph{Log Binomial.}
We evaluate the effect of using a log binomial distribution by studying the performance of \textit{ZoeD-M12-N} on NYU-Depth-v2 with log binomial and softmax probability heads. Consistent with \cite{unimodal-pmlr-v70-beckham17a}, we observe that using log binomial (REL = 0.075) instead of softmax (REL = 0.077) leads to about 2\% improvement. This highlights the importance of unimodal distributions for ordinal problems.

\begin{table}[]
\footnotesize
\setlength{\tabcolsep}{3pt} 
\centering
\begin{tabular}{c|cc|cc|cc}
\toprule
                 & \multicolumn{2}{c|}{\textbf{Labels required}} & \multicolumn{2}{c|}{\textbf{REL $\downarrow$}} & \multicolumn{2}{c}{\textbf{RMSE $\downarrow$}} \\

\textbf{Variant} & {Train}      & {Inference}      & {NYU}   & {KITTI} & {NYU}   & {KITTI} \\ \midrule
Labeled Router  & \cmark              & \cmark                  &      \underline{0.080}          &       \textbf{0.057} &  \underline{0.290} & \underline{2.452}   \\
Trained Router   & \cmark              & \xmark                  &       \textbf{0.077}    &    \textbf{0.057}       & \textbf{0.277} &  \textbf{ 2.362}  \\
Auto Router      & \xmark              & \xmark                  &      0.102          &  0.075  &  0.377  & {2.584}        \\ \bottomrule
\end{tabular}
\vspace{-6pt}
\caption{\textbf{Router variants.} The reported results are all based on \textit{ZoeD-M12-NK} evaluated on NYU Depth v2 and KITTI. Best results are in bold, second best are underlined.}
\label{tab:ablation-router}
\end{table}

\section{Conclusion}
\label{sec:conclusion}

Our proposed framework, \emph{ZoeDepth}, bridges the gap between relative and metric depth estimation. In the first stage, we pre-train an encoder-decoder architecture using relative depth on a collection of datasets. In the second stage, we add domain-specific heads based on our new metric bins module to the decoder and fine-tune the model on one or more datasets for metric depth prediction. Our proposed architecture decisively improves upon the state of the art for NYU Depth v2 (21\% in REL) and also significantly improves upon the state of the art in zero-shot transfer. We expect that defining more granular domains beyond indoor and outdoor, and fine-tuning on more metric datasets can improve our results further. In future work, we would like to investigate a mobile architecture version of ZoeDepth, \eg, for on-device photo editing, and extend our work to stereo-image depth estimation.

\clearpage
{\small
\bibliographystyle{ieee_fullname}
\bibliography{egbib}
}
\clearpage

\appendix
\section{Appendix}
\subsection{Datasets Overview} 
\label{sec:datasets-overview}
We begin by providing a detailed overview of the properties of the datasets used for metric depth fine-tuning and evaluation of the new ZoeDepth architecture (see Fig.~2 in the main paper) in \Cref{tab:datasets-overview}. These datasets consist of NYU Depth v2~\cite{Silberman2012} and  KITTI~\cite{Menze_2015_CVPR} used for metric depth fine-tuning as well as respectively four in- and outdoor datasets to test for generalization performance (see Sec.~4.1 of the main paper). The indoor datasets consist of SUN RGB-D~\cite{Song2015_sunrgbd}, the iBims benchmark~\cite{koch2019}, DIODE Indoor~\cite{diode_dataset} and HyperSim ~\cite{roberts:2021}. For the outdoor datasets, we use DDAD~\cite{packnet}, DIML Outdoor~\cite{kim2018deep}, DIODE Outdoor~\cite{diode_dataset} and Virtual KITTI~2~\cite{cabon2020vkitti2}.

All ZoeDepth architectures and prior works are evaluated by resizing the input to the training resolution. Zoe-*-N, Zoe-*-NK, and Zoe-*-K models are trained at resolutions $384\times512$, $384\times512$ and $384\times768$ respectively. Predictions are resized to original ground truth resolution before evaluation.

\begin{table*}[!htb]
\centering
\setlength{\tabcolsep}{4pt} 
\begin{tabular}{l|c|c|c|c|c|c|c|c} 
    \toprule
     &  &  & Seen in & \# Train & \# Eval & \multicolumn{2}{c|}{Eval Depth [m]} & Crop\\
    Dataset & Domain & Type & Training? & Samples & Samples & Min & Max & Method\\
    \hline
    NYU Depth v2~\cite{Silberman2012} & Indoor & Real & \checkmark & 24k~\cite{bts_lee2019big} & 654 & 1e-3 & 10  & Eigen\\
    SUN RGB-D~\cite{Song2015_sunrgbd} & Indoor & Real & \xmark & - & 5050 & 1e-3 & 8  & Eigen\\
    iBims-1~\cite{koch2019} & Indoor & Real & \xmark & - & 100 & 1e-3 & 10  & Eigen\\
    DIODE Indoor~\cite{diode_dataset} & Indoor & Real & \xmark & - & 325 & 1e-3 & 10  & Eigen\\
    HyperSim~\cite{roberts:2021} & Indoor & Synthetic & \xmark & - & 7690 & 1e-3 & 80  & Eigen\\
    \hline
    KITTI~\cite{Menze_2015_CVPR} & Outdoor & Real & \checkmark & 26k~\cite{bts_lee2019big} & 697 & 1e-3 & 80  & Garg$\ddagger$\\
    Virtual KITTI 2~\cite{cabon2020vkitti2} & Outdoor & Synthetic & \xmark & - & 1701 & 1e-3 & 80  & Garg$\ddagger$\\
    DDAD~\cite{packnet} & Outdoor & Real & \xmark & - & 3950 & 1e-3 & 80  & Garg\\
    DIML Outdoor~\cite{kim2018deep} & Outdoor & Real & \xmark & - & 500 & 1e-3 & 80  & Garg\\
    DIODE Outdoor~\cite{diode_dataset} & Outdoor & Real & \xmark & - & 446 & 1e-3 & 80  & Garg\\
    \bottomrule
    \end{tabular}
\caption{Overview of datasets used in metric depth fine-tuning and evaluation of ZoeDepth architectures. For demonstrating zero-shot transfer, we evaluate across a total of 13165 indoor samples and 6597 outdoor samples. While HyperSim is predominantly an indoor dataset, there are several samples exhibiting depth ranges exceeding 10 m, so we relax the maximum evaluation depth up to 80 m. $\ddagger$ : To follow prior works \cite{yuan2022new, bhat2021adabins}, we crop the sample and then use scaled Garg crop for evaluation. We verify the transforms by reproducing results obtained by using respective pre-trained checkpoints provided by prior works.}
\label{tab:datasets-overview}
\end{table*}

\subsection{Training Details} 
In ~\Cref{tab:model-def}, we show various training strategies (see Section 3.3 of the main paper) applied to the ZoeDepth architecture. The training strategies differ by the datasets used for relative depth pre-training of the MiDaS~\cite{Ranftl2020MiDaS} encoder-decoder, the datasets employed for metric depth fine-tuning in ZoeDepth and the number of used metric heads. Each combination of these options provided in \cref{tab:model-def} defines a different model of ZoeDepth. Results for these combinations are shown in \Cref{sec:additional-results}.

\begin{table*}[!htb]
\centering
\begin{tabular}{l|c|c|c} 
    \toprule
    Model       & Relative depth pre-training & Metric depth fine-tuning & \# metric heads \\ 
    \hline
    ZoeD-X-N        & \xmark            & NYU     & 1   \\
    ZoeD-N-N        & NYU               & NYU     & 1   \\
    ZoeD-NK-N       & NYU+KITTI         & NYU     & 1   \\
    ZoeD-M12-N      & M12               & NYU     & 1   \\
    \midrule
    ZoeD-X-K        & \xmark            & KITTI     & 1   \\
    ZoeD-K-K        & KITTI             & KITTI     & 1   \\
    ZoeD-NK-K       & NYU+KITTI         & KITTI     & 1   \\
    ZoeD-M12-K      & M12               & KITTI     & 1   \\
    \midrule
    ZoeD-NK-NK$^\dagger$                & NYU+KITTI         & NYU+KITTI        & 1  \\
    ZoeD-M12-NK$^\dagger$               & M12               & NYU+KITTI        & 1  \\
    \midrule
    ZoeD-NK-NK            & NYU+KITTI         & NYU+KITTI        & 2  \\
    ZoeD-M12-NK           & M12               & NYU+KITTI        & 2  \\
    \bottomrule
    \end{tabular}
\caption{Models are named according to the following convention: ZoeD-{RDPT}-{MFT}, where ZoeD is the abbreviation for ZoeDepth, RDPT denotes the datasets used for relative depth pre-training and MFT denotes the datasets used for metric depth fine-tuning. Models with an X do not use a relative depth pre-training. The collection M12 contains the datasets HRWSI~\cite{xian2020structure}, BlendedMVS~\cite{yao2020blendedmvs}, ReDWeb~\cite{xian2018monocular}, DIML-Indoor~\cite{kim2018deep}, 3D Movies~\cite{Ranftl2020MiDaS}, MegaDepth~\cite{MDLi18}, WSVD~\cite{wang2019web}, TartanAir~\cite{wang2020tartanair}, ApolloScape~\cite{huang2019apolloscape}, IRS~\cite{wang2019irs}, KITTI~(K)~\cite{Menze_2015_CVPR} and NYU Depth v2~(N)~\cite{Silberman2012}.}
\label{tab:model-def}
\end{table*}

\subsection{Detailed Results} 
\label{sec:additional-results}
In Tables~3 and~4 of the main paper, we have provided quantitative results for zero-shot transfer to the four in- and outdoor datasets unseen during training, which are mentioned in \Cref{sec:datasets-overview}. These results are supplemented by the threshold accuracies $\delta_2$ and $\delta_3$ as well as the average $\log_{10}$ error in \Cref{tab:sunrgbd-full,tab:ibims-full,tab:diode-indoor-full,tab:hypersim-full,tab:vkitti2-full,tab:ddad-full,tab:diml-outdoor-full,tab:diode-outdoor-full}. Also, while the main paper only shows our models ZoeD-X-K, ZoeD-M12-K and ZoeD-M12-NK, \Cref{tab:sunrgbd-full,tab:ibims-full,tab:diode-indoor-full,tab:hypersim-full,tab:vkitti2-full,tab:ddad-full,tab:diml-outdoor-full,tab:diode-outdoor-full} contain the additional model ZoeD-NK-N. This model uses only the dataset combination of NYU Depth v2 (N)~\cite{Silberman2012} and KITTI (K)~\cite{Menze_2015_CVPR} for the relative depth pre-training.

\begin{figure*}[htb]
    \begin{overpic}[width=\textwidth]{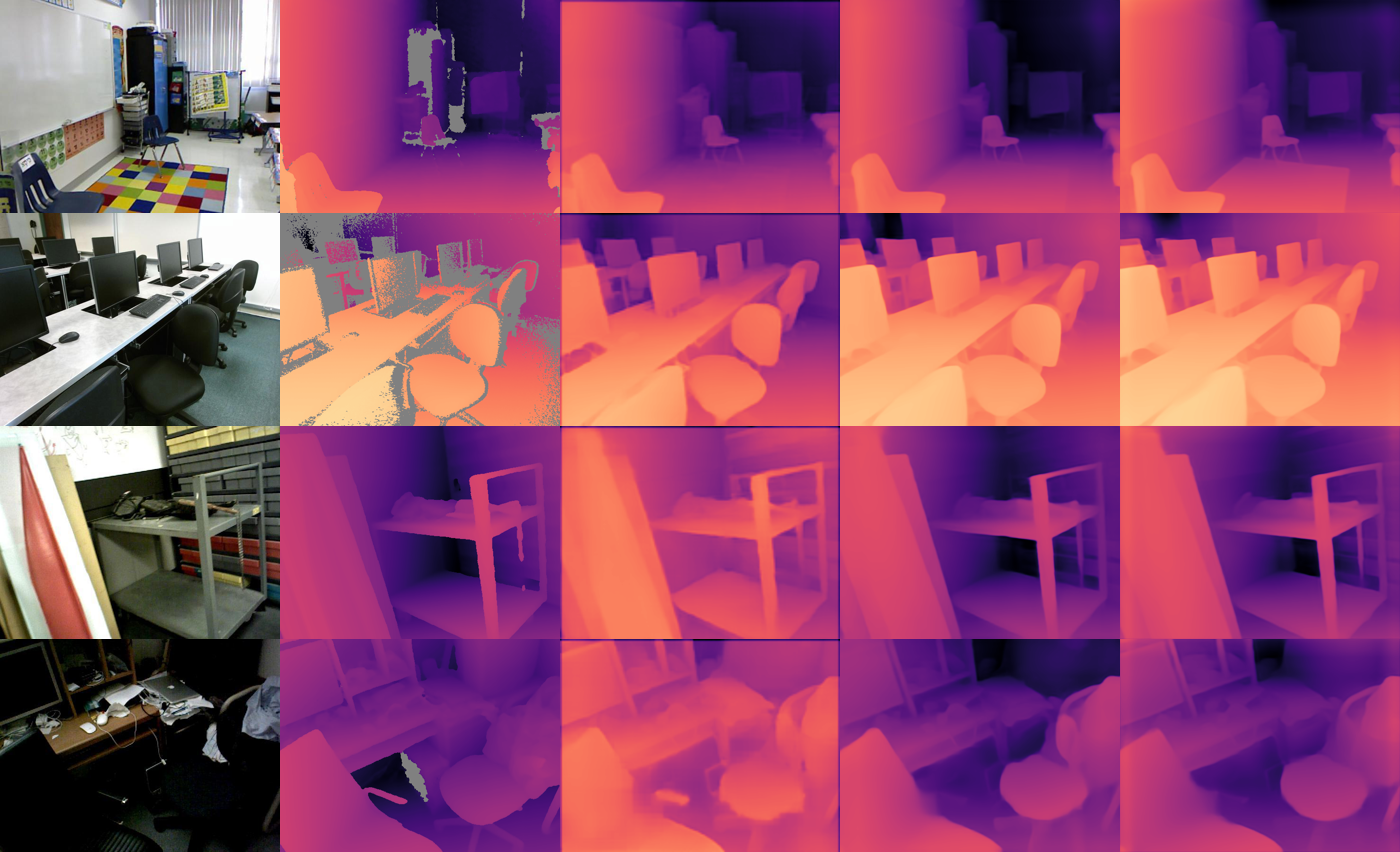}
    \put(8,-2){RGB}
    \put(24,-2){Ground Truth}
    \put(45,-2){NeWCRF~\cite{yuan2022new}}
    \put(64,-2){Zoe-M12-NK}
    \put(85,-2){Zoe-M12-N}
    \end{overpic}
    \\
    \caption{Zero-shot transfer to the SUN RGB-D dataset~\cite{Song2015_sunrgbd}. Invalid regions are indicated in gray.}
    \label{fig:sunrgbd}
\end{figure*}

\begin{figure*}[htb]
    \begin{overpic}[width=\textwidth]{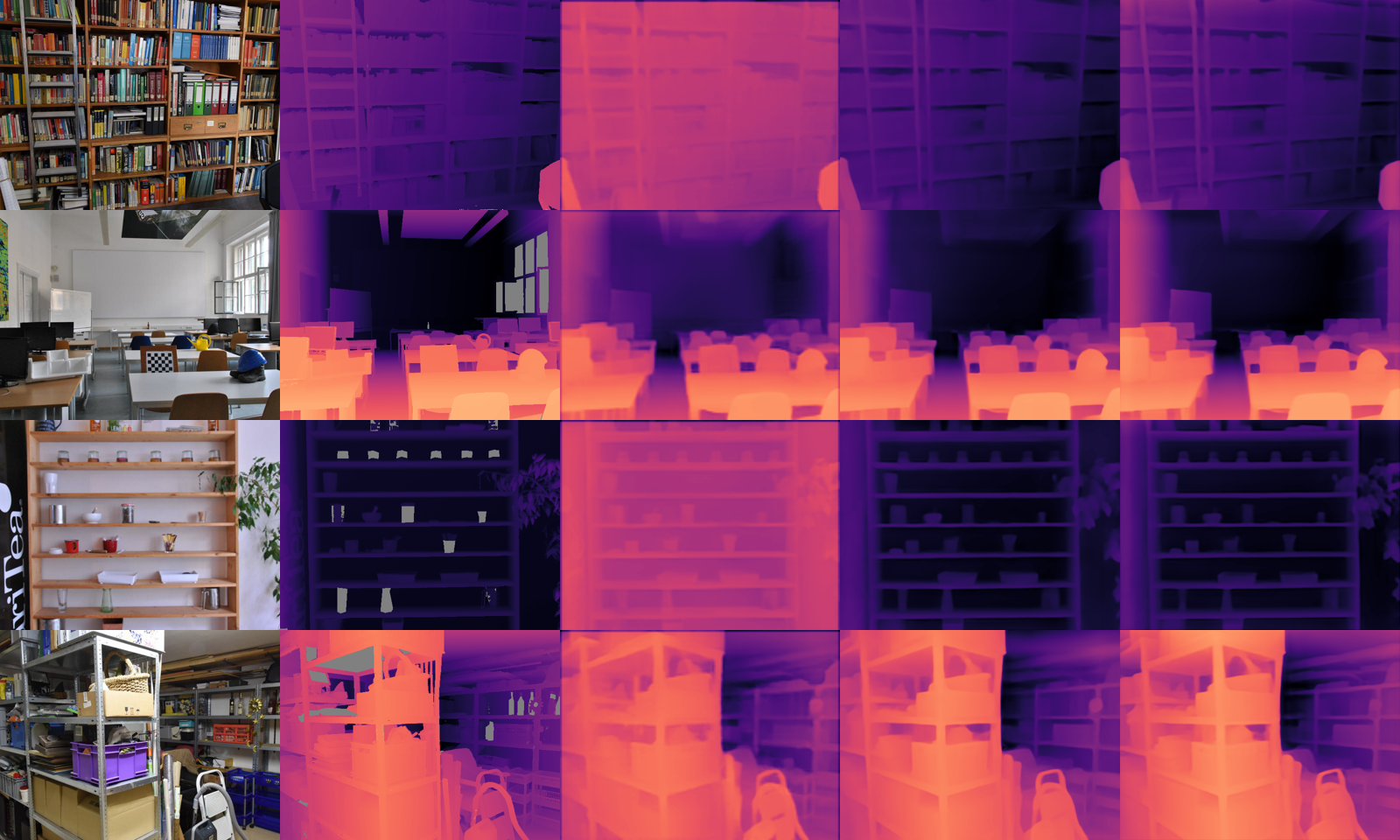}
    \put(8,-2){RGB}
    \put(24,-2){Ground Truth}
    \put(45,-2){NeWCRF~\cite{yuan2022new}}
    \put(64,-2){Zoe-M12-NK}
    \put(85,-2){Zoe-M12-N}
    \end{overpic}
    \\
    \caption{Zero-shot transfer to the iBims-1 benchmark~\cite{koch2019}. Invalid regions are indicated in gray.}
    \label{fig:ibims}
\end{figure*}

\begin{figure*}[htb]
    \begin{overpic}[width=\textwidth]{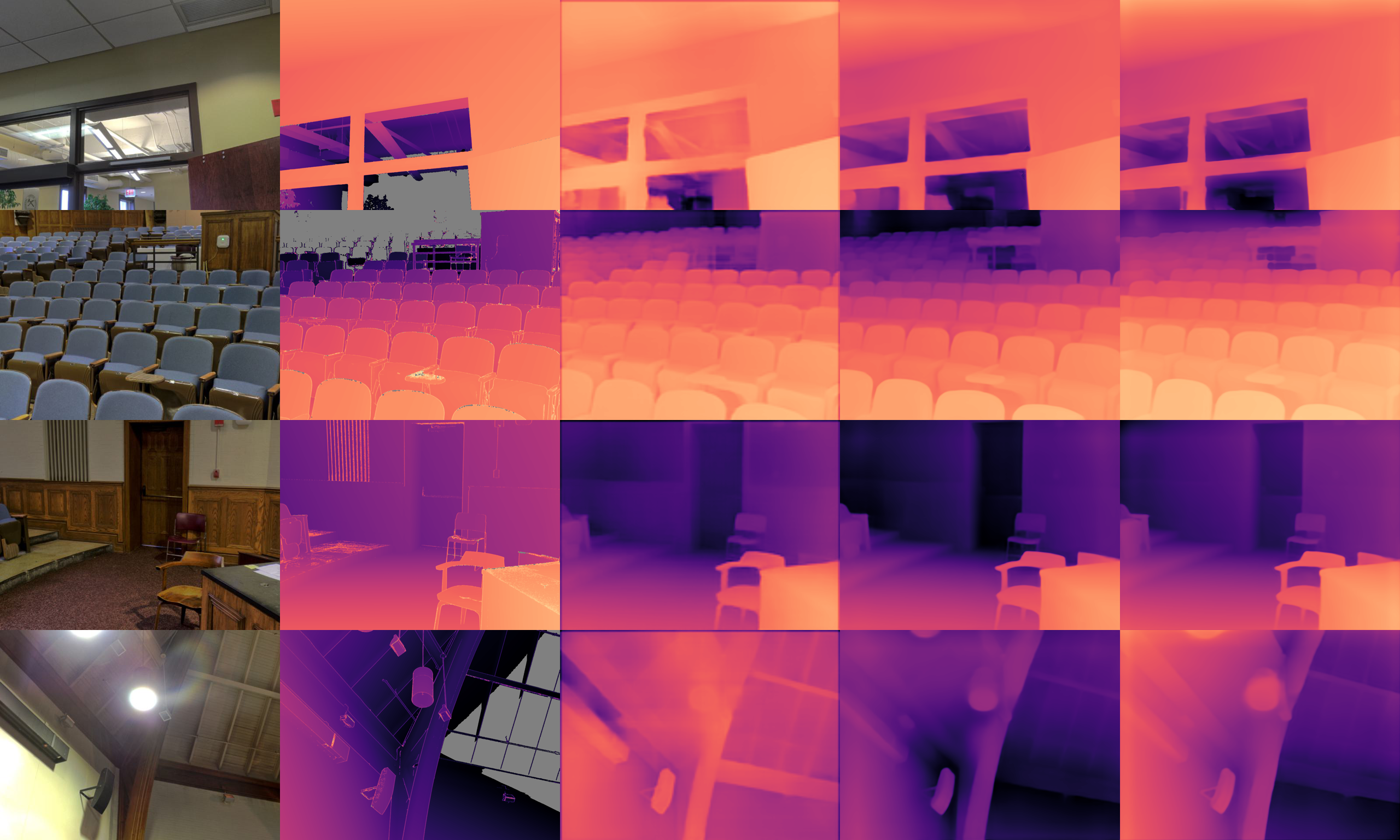}
    \put(8,-2){RGB}
    \put(24,-2){Ground Truth}
    \put(45,-2){NeWCRF~\cite{yuan2022new}}
    \put(64,-2){Zoe-M12-NK}
    \put(85,-2){Zoe-M12-N}
    \end{overpic}
    \\
    \caption{Zero-shot transfer to the DIODE Indoor dataset~\cite{diode_dataset}. Invalid regions are indicated in gray.}
    \label{fig:diode-indoor}
\end{figure*}

\begin{figure*}[htb]
    \begin{overpic}[width=\textwidth]{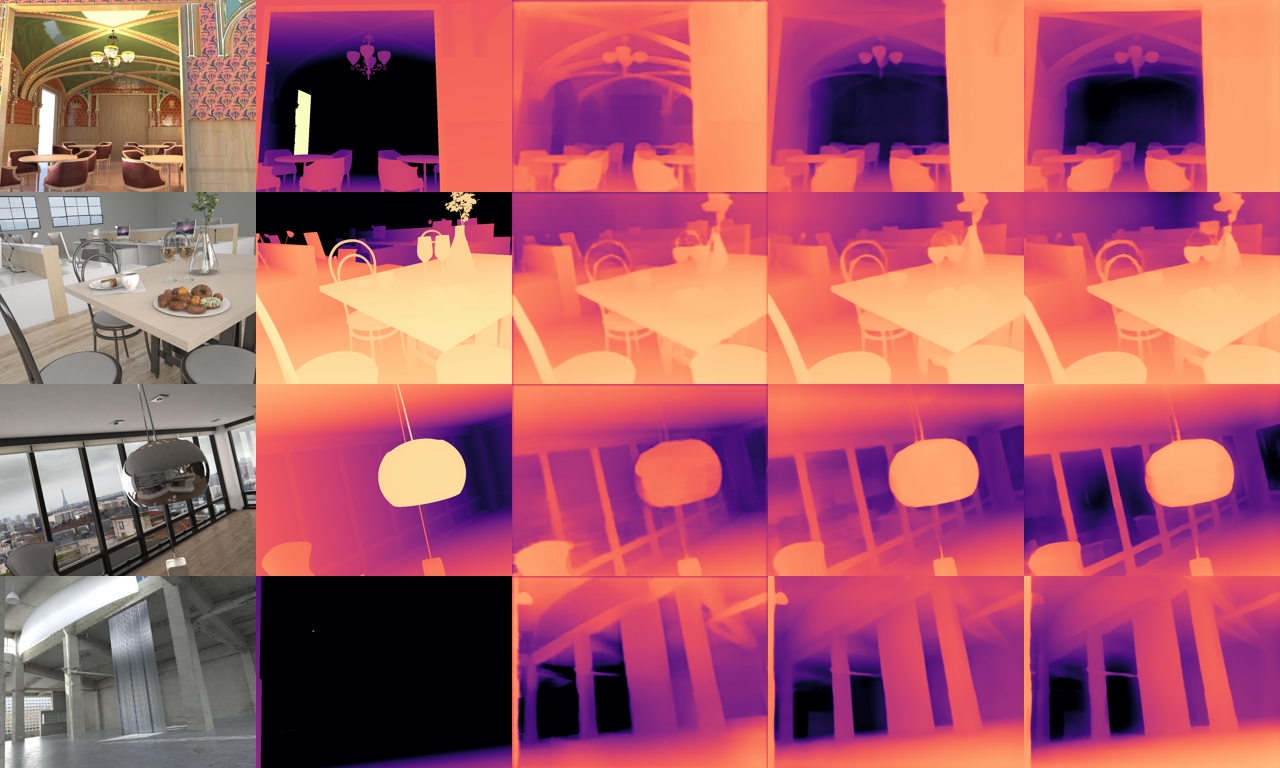}
    \put(8,-2){RGB}
    \put(24,-2){Ground Truth}
    \put(45,-2){NeWCRF~\cite{yuan2022new}}
    \put(64,-2){Zoe-M12-NK}
    \put(85,-2){Zoe-M12-N}
    \end{overpic}
    \\
    \caption{Zero-shot transfer to the HyperSim dataset~\cite{roberts:2021}.}
    \label{fig:hypersim-full}
\end{figure*}

\begin{figure*}[htb]
    \begin{overpic}[width=\textwidth]{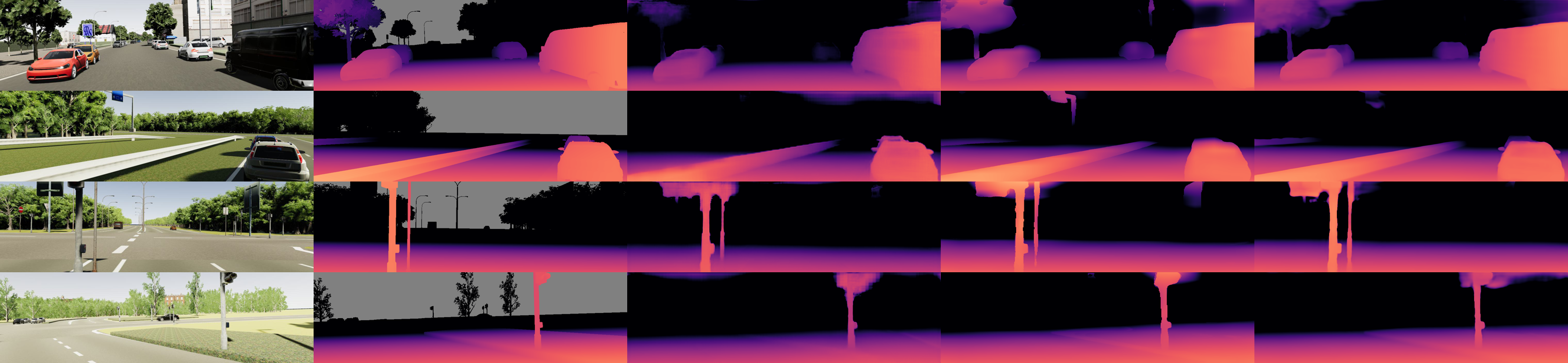}
    \put(8,-2){RGB}
    \put(24,-2){Ground Truth}
    \put(45,-2){NeWCRF~\cite{yuan2022new}}
    \put(64,-2){Zoe-M12-NK}
    \put(85,-2){Zoe-M12-K}
    \end{overpic}
    \\
    \caption{Zero-shot transfer to the Virtual KITTI 2 dataset~\cite{cabon2020vkitti2}. Invalid regions are indicated in gray.}
    \label{fig:vkitti2-full}
\end{figure*}

\begin{figure*}[htb]
    \begin{overpic}[width=\textwidth]{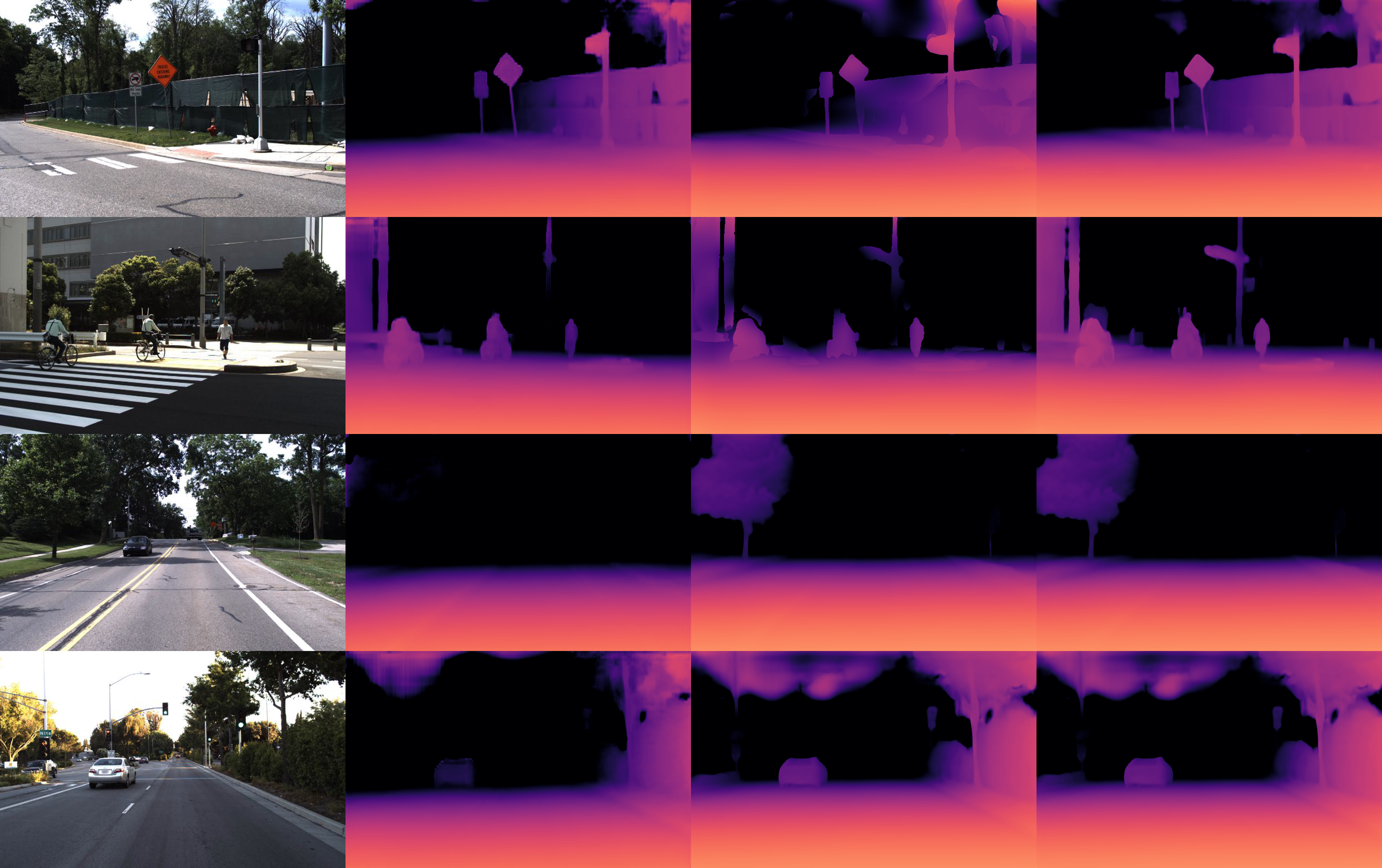}
    \put(8,-2){RGB}
    \put(33,-2){NeWCRF~\cite{yuan2022new}}
    \put(58,-2){Zoe-M12-NK}
    \put(80,-2){Zoe-M12-K}
    \end{overpic}
    \\
    \caption{Zero-shot transfer to the DDAD dataset~\cite{packnet}. Ground truth depth is too sparse to visualize here.}
    \label{fig:ddad}
\end{figure*}

\begin{figure*}[htb]
    \begin{overpic}[width=\textwidth]{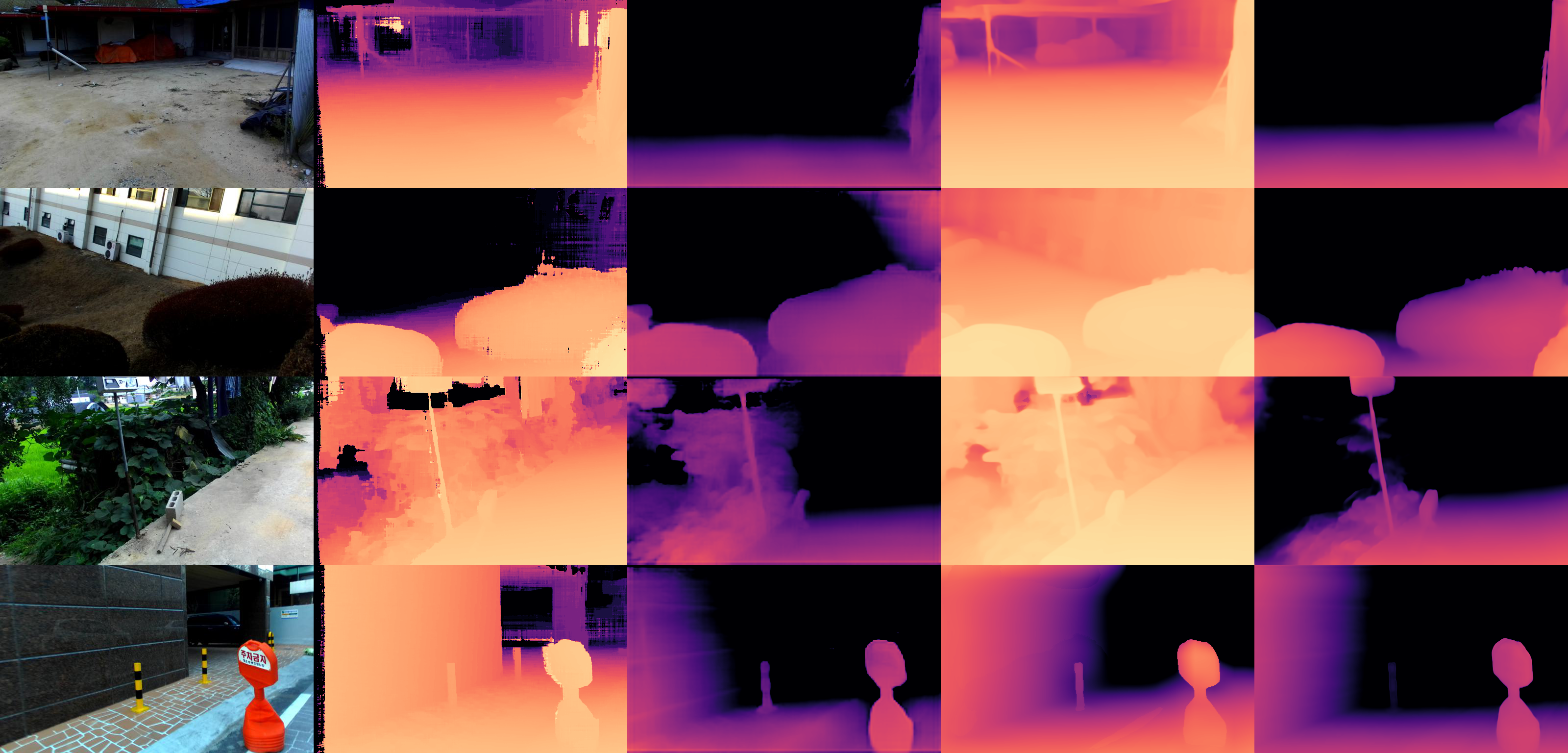}
    \put(8,-2){RGB}
    \put(24,-2){Ground Truth}
    \put(45,-2){NeWCRF~\cite{yuan2022new}}
    \put(64,-2){Zoe-M12-NK}
    \put(85,-2){Zoe-M12-K}
    \end{overpic}
    \\
    \caption{Zero-shot transfer to the DIML Outdoor dataset~\cite{kim2018deep}.}
    \label{fig:diml-outdoor}
\end{figure*}

\begin{figure*}[htb]
    \begin{overpic}[width=\textwidth]{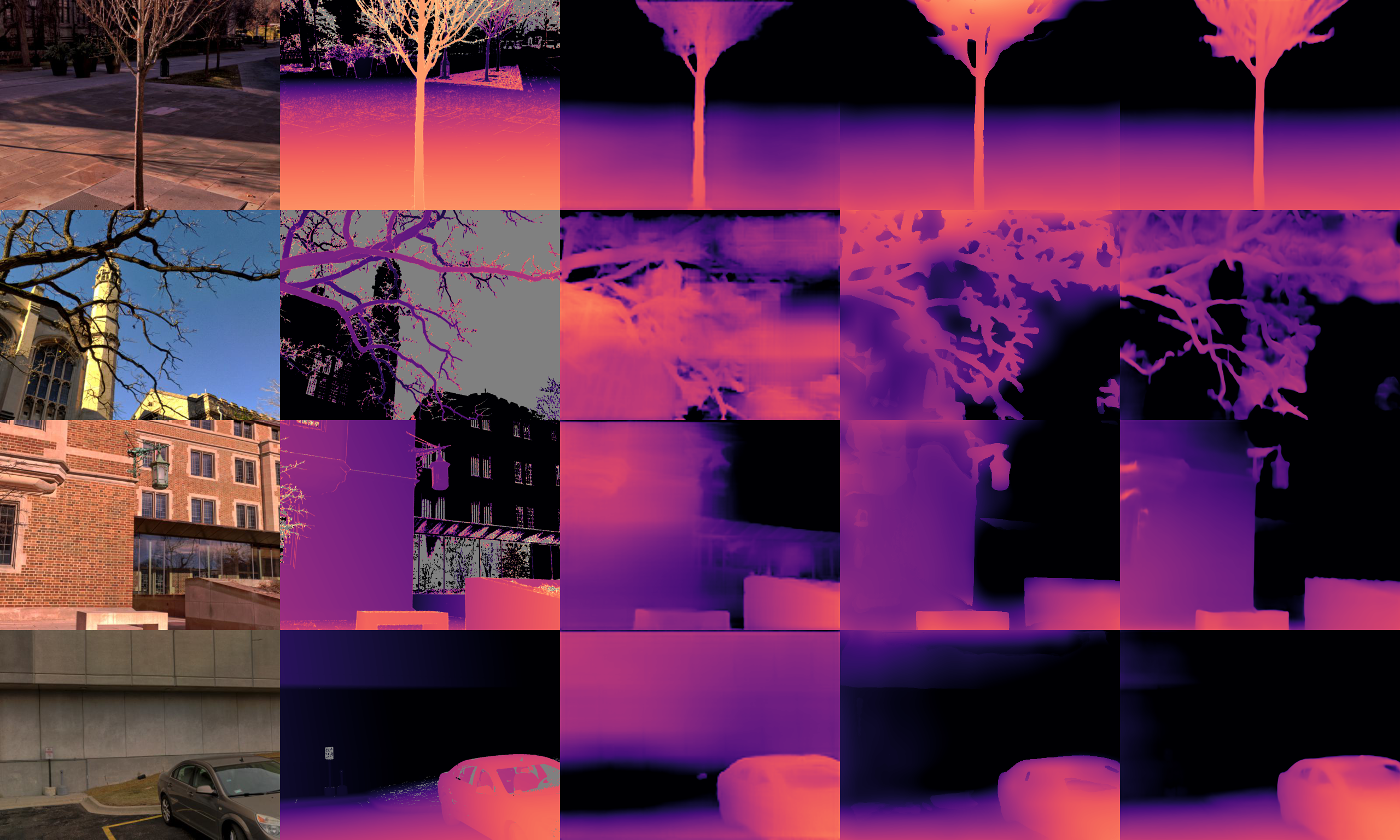}
    \put(8,-2){RGB}
    \put(24,-2){Ground Truth}
    \put(45,-2){NeWCRF~\cite{yuan2022new}}
    \put(64,-2){Zoe-M12-NK}
    \put(85,-2){Zoe-M12-K}
    \end{overpic}
    \\
    \caption{Zero-shot transfer to the DIODE Outdoor dataset~\cite{diode_dataset}. Invalid regions are indicated in gray.}
    \label{fig:diode-outdoor}
\end{figure*}

\Cref{fig:sunrgbd,fig:ibims,fig:diode-indoor,fig:hypersim-full,fig:vkitti2-full,fig:ddad,fig:diml-outdoor,fig:diode-outdoor} show metric depth maps computed with our ZoeDepth architecture for various example images of the in- and outdoor datasets described in \Cref{sec:datasets-overview}.

For the indoor datasets, NeWCRF shows a tendency to underestimate the depth, \eg, the relatively bright images of NeWCRF in rows 3 and 4 of \Cref{fig:sunrgbd} as well as rows 1 and 3 of \Cref{fig:ibims}. Our models Zoe-M12-NK and Zoe-M12-N are much closer to the ground truth. Only row 4 of \Cref{fig:hypersim-full} is an exception where our models share this behavior with NeWCRF. 

For the outdoor datasets, rows 1 to 3 of \Cref{fig:diml-outdoor} clearly demonstrate the advantage of our model ZoeD-M12-NK with respect to NeWCRF. As explained in Section~5.3 of the main paper, ZoeD-M12-NK is not only fine-tuned on the outdoor dataset KITTI but also on the indoor dataset NYU Depth v2, which better reflects the low depth values observable in the RGB and ground truth images of \Cref{fig:diml-outdoor}. The improved sharpness in predictions from our models when compared to NeWCRF, as mentioned in the caption of Figure 4 of the main paper, continues to hold across all 8 indoor and outdoor datasets.

\begin{table}[htb]
\centering
\setlength{\tabcolsep}{3pt} 
\small
\begin{tabular}{@{}lcccccc@{}}
\toprule
Method & $\delta_1\uparrow$ & $\delta_2\uparrow$ & $\delta_3\uparrow$ & REL~$\downarrow$ & RMSE~$\downarrow$ & $\log_{10}\downarrow$\\ \midrule
BTS~\cite{bts_lee2019big}               & 0.740     & 0.933     & 0.980     & 0.172     & 0.515     & 0.075     \\ 
AdaBins~\cite{bhat2021adabins}          & 0.771     & 0.944     & 0.983     & 0.159     & 0.476     & 0.068     \\ 
LocalBins~\cite{bhat2022localbins}      & {0.777}   & {0.949}   & {0.985}   & {0.156}   & {0.470}   &  {0.067}  \\
NeWCRF~\cite{yuan2022new}               & {0.798}   & {0.967}   & {0.992}   & {0.151}   & {0.424}   &  {0.064}  \\ 

\midrule
\textbf{ZoeD-X-N}       & \underline{0.857} & \underline{0.979} & \textbf{0.995} & {0.124} &  0.363 & {0.054} \\ 
\textbf{ZoeD-NK-N}      & \underline{0.857} & 0.978 & \underline{0.994} & 0.125 &  \underline{0.360}  & {0.054} \\
\textbf{ZoeD-M12-N}     & \textbf{0.864} &            \textbf{0.982} &            \textbf{0.995} &                \textbf{ 0.119} &              \textbf{0.346} &                \textbf{0.052} \\
\textbf{ZoeD-M12-NK}    & 0.856 &            \underline{0.979} &           \textbf{ 0.995} &                 \underline{0.123} &              \underline{0.356} &                \underline{0.053} \\
\bottomrule
\end{tabular}
\caption{Zero-shot transfer to the SUN RGB-D dataset \cite{Song2015_sunrgbd}. Best results are in bold, second best are underlined.}
\label{tab:sunrgbd-full}
\end{table}

\begin{table}[htb]
\centering
\setlength{\tabcolsep}{3pt} 
\small
\begin{tabular}{@{}lcccccc@{}}
\toprule
Method & $\delta_1\uparrow$ & $\delta_2\uparrow$ & $\delta_3\uparrow$ & REL~$\downarrow$ & RMSE~$\downarrow$ & $\log_{10}\downarrow$\\ \midrule
BTS~\cite{bts_lee2019big}               & 0.538     & 0.863     & 0.948     & 0.231     & 0.919     & 0.112    \\ 
AdaBins~\cite{bhat2021adabins}          & 0.555     & 0.873     & 0.960     & 0.212     & 0.901     & 0.107    \\ 
LocalBins~\cite{bhat2022localbins}      & 0.558     & 0.877     & 0.966     & 0.211     & 0.880     & 0.104     \\
NeWCRF~\cite{yuan2022new}               & 0.548     & 0.884     & 0.979     & 0.206     & 0.861     & 0.102     \\ 
\midrule
\textbf{ZoeD-X-N}       & \underline{0.668} & \underline{0.944} & \underline{0.983} &     0.173 &  \underline{0.730}  &    \underline{0.084} \\ 
\textbf{ZoeD-NK-N}      &  \textbf{0.671} & 0.939 & \underline{0.983} &     \underline{0.172} &  0.735 &    \underline{0.084} \\
\textbf{ZoeD-M12-N}     &  0.658 &         \textbf{ 0.947} &         \textbf{ 0.985} &               \textbf{0.169} &            \textbf{0.711} &              \textbf{0.083} \\
\textbf{ZoeD-M12-NK}   & 0.615 &          0.928 &          0.982 &               0.186 &            0.777 &              0.092 \\
\bottomrule
\end{tabular}
\caption{Zero-shot transfer to the iBims-1 benchmark~\cite{koch2019}. Best results are in bold, second best are underlined.}
\label{tab:ibims-full}
\end{table}

\begin{table}[htb]
\centering
\setlength{\tabcolsep}{3pt} 
\small
\begin{tabular}{@{}lcccccc@{}}
\toprule
Method & $\delta_1\uparrow$ & $\delta_2\uparrow$ & $\delta_3\uparrow$ & REL~$\downarrow$ & RMSE~$\downarrow$ & $\log_{10}\downarrow$\\ \midrule
BTS~\cite{bts_lee2019big}               & 0.210     & 0.478     & 0.699     & 0.418    & 1.905     & 0.250    \\ 
AdaBins~\cite{bhat2021adabins}          & 0.174     & 0.438     & 0.658     & 0.443    & 1.963     & 0.270    \\ 
LocalBins~\cite{bhat2022localbins}      & 0.229 & 0.520  & 0.718 &     0.412 &  1.853 &    0.246     \\
NeWCRF~\cite{yuan2022new}               & 0.187     & 0.498     & 0.748     & 0.404    & 1.867     & 0.241     \\ 
\midrule
\textbf{ZoeD-X-N}       & \textbf{0.400}   & \textbf{0.704} & 0.808 &     \textbf{0.324} &  \textbf{1.581} &    \textbf{0.181} \\ 
\textbf{ZoeD-NK-N}      & 0.365 & \underline{0.696} & \underline{0.819} &     0.335 &  1.604 &    0.188 \\
\textbf{ZoeD-M12-N}  & 0.376 & \underline{0.696} & \textbf{0.822} &    \underline{0.327} & \underline{1.588} &   0.186 \\
\textbf{ZoeD-M12-NK}    & \underline{0.386} & 0.695 & 0.807 & 0.331 &   1.598 &     \underline{0.185} \\
\bottomrule
\end{tabular}
\caption{Zero-shot transfer to the DIODE Indoor dataset~\cite{diode_dataset}. Best results are in bold, second best are underlined.}
\label{tab:diode-indoor-full}
\end{table}

\begin{table}[htb]
\centering
\setlength{\tabcolsep}{3pt} 
\small
\begin{tabular}{@{}lcccccc@{}}
\toprule
Method & $\delta_1\uparrow$ & $\delta_2\uparrow$ & $\delta_3\uparrow$ & REL~$\downarrow$ & RMSE~$\downarrow$ & $\log_{10}\downarrow$\\ \midrule
BTS~\cite{bts_lee2019big}               & 0.225     & 0.419     & 0.582     & 0.476     & 6.404     & 0.329    \\ 
AdaBins~\cite{bhat2021adabins}          & 0.221     & 0.410     & 0.568     & 0.483     & 6.546     & 0.345    \\ 
LocalBins~\cite{bhat2022localbins}      & 0.234 & 0.432 & 0.594 &     0.468 &  6.362 &    0.320     \\
NeWCRF~\cite{yuan2022new}               & 0.255     & 0.464     & 0.638     & 0.442     & 6.017     & 0.283     \\ 
\midrule
\textbf{ZoeD-X-N}       & 0.284 & 0.502 & 0.692 &     0.421 &  5.889 &    0.267 \\ 
\textbf{ZoeD-NK-N}      &  \underline{0.291} & \textbf{0.519} & \underline{0.700}   &     \underline{0.414} &  5.838 &    \underline{0.260} \\
\textbf{ZoeD-M12-N}     & \textbf{0.292} & \underline{0.514} & \textbf{0.706} & \textbf{0.410} & \textbf{5.771} &   \textbf{0.257} \\
\textbf{ZoeD-M12-NK}    &  0.274 & 0.494 & 0.696 & 0.419 & 5.830 &   0.262 \\
\bottomrule
\end{tabular}
\caption{Zero-shot transfer to the HyperSim dataset~\cite{roberts:2021}. Best results are in bold, second best are underlined.}
\label{tab:hypersim-full}
\end{table}

\begin{table}[htb]
\centering
\setlength{\tabcolsep}{3pt} 
\small
\begin{tabular}{@{}lcccccc@{}}
\toprule
Method & $\delta_1\uparrow$ & $\delta_2\uparrow$ & $\delta_3\uparrow$ & REL~$\downarrow$ & RMSE~$\downarrow$ & $\log_{10}\downarrow$\\ \midrule
BTS~\cite{bts_lee2019big}               & 0.831 & 0.948 & 0.982 &     0.115 &  5.368 &    0.054   \\ 
AdaBins~\cite{bhat2021adabins}          & 0.826 & 0.947 & 0.98  &     0.122 &  5.42  &    0.057    \\ 
LocalBins~\cite{bhat2022localbins}      & 0.810  & 0.94  & 0.978 &     0.127 &  5.981 &    0.061    \\
NeWCRF~\cite{yuan2022new}               & 0.829 & 0.951 & 0.984 &     0.117 &  5.691 &    0.056     \\ 
\midrule
\textbf{ZoeD-X-K}       & 0.837 & 0.965 & \underline{0.991} &     0.112 &  5.338 &    0.053 \\ 
\textbf{ZoeD-NK-K}      & \underline{0.855} & \underline{0.970}  & \textbf{0.992} &     \underline{0.101} &  {5.102} &    \underline{0.048} \\
\textbf{ZoeD-M12-K}     & \textbf{0.864} & \textbf{0.973} & \textbf{0.992} &     \textbf{0.100}   &  \textbf{4.974} &    \textbf{0.046} \\
\textbf{ZoeD-M12-NK}    & 0.850 & 0.965 & \underline{0.991} & 0.105 & \underline{5.095} & 0.050 \\
\bottomrule
\end{tabular}
\caption{Zero-shot transfer to the Virtual KITTI 2 dataset~\cite{cabon2020vkitti2}. Best results are in bold, second best are underlined.}
\label{tab:vkitti2-full}
\end{table}

\begin{table}[htb]
\centering
\setlength{\tabcolsep}{3pt} 
\small
\begin{tabular}{@{}lcccccc@{}}
\toprule

Method & $\delta_1\uparrow$ & $\delta_2\uparrow$ & $\delta_3\uparrow$ & REL~$\downarrow$ & RMSE~$\downarrow$ & $\log_{10}\downarrow$\\ \midrule
BTS~\cite{bts_lee2019big}               & 0.805     & 0.945     & 0.982     & 0.147     & 7.550     & 0.067    \\ 
AdaBins~\cite{bhat2021adabins}          & 0.766     & 0.918     & 0.972     & 0.154     & 8.560     & 0.074    \\ 
LocalBins~\cite{bhat2022localbins}      & 0.777     & 0.930     & 0.978     & 0.151     & 8.139     & 0.071     \\
NeWCRF~\cite{yuan2022new}               & \textbf{0.874}     & \textbf{0.974}     & \textbf{0.991}     & \textbf{0.119}     & \textbf{6.183}     & \textbf{0.054}     \\ 
\midrule
\textbf{ZoeD-X-K}       &  0.790  & 0.95  & 0.985 &     0.137 &  7.734 &    0.066 \\ 
\textbf{ZoeD-NK-K}      & 0.824 & 0.957 & 0.987 &     0.134 &  7.249 &    0.062 \\
\textbf{ZoeD-M12-K}     & \underline{0.835} & \underline{0.962} & \underline{0.988} &     \underline{0.129} &  \underline{7.108} &    \underline{0.060} \\
\textbf{ZoeD-M12-NK}    & 0.824 & 0.951 &  0.980 & 0.138 & 7.225 & 0.066 \\
\bottomrule
\end{tabular}
\caption{Zero-shot transfer to the DDAD dataset~\cite{packnet}. Best results are in bold, second best are underlined.}
\label{tab:ddad-full}
\end{table}

\begin{table}[htb]
\centering
\setlength{\tabcolsep}{3pt} 
\small
\begin{tabular}{@{}lcccccc@{}}
\toprule
Method & $\delta_1\uparrow$ & $\delta_2\uparrow$ & $\delta_3\uparrow$ & REL~$\downarrow$ & RMSE~$\downarrow$ & $\log_{10}\downarrow$\\ \midrule
BTS~\cite{bts_lee2019big}               & \underline{0.016}     & 0.042     & 0.123     & 1.785     & \underline{5.908}     & \underline{0.428}    \\ 
AdaBins~\cite{bhat2021adabins}          & 0.013     & 0.038     & 0.107     & 1.941     & 6.272     & 0.451    \\ 
LocalBins~\cite{bhat2022localbins}      & \underline{0.016} & \underline{0.044} & \underline{0.124} &     1.82  &  6.706 &    0.434     \\
NeWCRF~\cite{yuan2022new}               & 0.010     & 0.032     & 0.094     & 1.918     & 6.283     & 0.449     \\ 
\midrule
\textbf{ZoeD-X-K}       & 0.005 & 0.022 & 0.096 &     \underline{1.756} &  6.180  &  0.429 \\ 
\textbf{ZoeD-NK-K}      & 0.004 & 0.012 & 0.047 &     2.068 &  7.432  &  0.473 \\ 
\textbf{ZoeD-M12-K}     & 0.003 & 0.010 & 0.048 &     1.921 &  6.978  &  0.455 \\
\textbf{ZoeD-M12-NK}    & \textbf{0.292} &\textbf{ 0.562} & \textbf{0.697} &      \textbf{0.641} &    \textbf{3.610} &     \textbf{0.213} \\
\bottomrule
\end{tabular}
\caption{Zero-shot transfer to the DIML Outdoor dataset~\cite{kim2018deep}. Best results are in bold, second best are underlined.}
\label{tab:diml-outdoor-full}
\end{table}

\begin{table}[htb]
\centering
\setlength{\tabcolsep}{3pt} 
\small
\begin{tabular}{@{}lcccccc@{}}
\toprule
Method & $\delta_1\uparrow$ & $\delta_2\uparrow$ & $\delta_3\uparrow$ & REL~$\downarrow$ & RMSE~$\downarrow$ & $\log_{10}\downarrow$\\ \midrule
BTS~\cite{bts_lee2019big}               & 0.171     & 0.347     & 0.526     & 0.837     & 10.48     & 0.334    \\ 
AdaBins~\cite{bhat2021adabins}          & 0.161     & 0.329     & 0.529     & 0.863     & 10.35     & 0.318    \\ 
LocalBins~\cite{bhat2022localbins}      & 0.170     & 0.336     & 0.531     & \underline{0.821}     & 10.273    & 0.329     \\
NeWCRF~\cite{yuan2022new}               & 0.176     & 0.369     & 0.588     & 0.854     & 9.228     & 0.283     \\ 
\midrule
\textbf{ZoeD-X-K}       & \underline{0.242} & 0.485 & 0.744 & \textbf{0.799} &  7.806 &    0.219 \\ 
\textbf{ZoeD-NK-K}      & 0.241 & \underline{0.505} & \underline{0.759} & 0.892 &  \underline{7.489} &    \underline{0.216} \\
\textbf{ZoeD-M12-K}     & \textbf{0.269} & \textbf{0.563} & \textbf{0.816} & 0.852 &  \textbf{6.898} &    \textbf{0.198} \\
\textbf{ZoeD-M12-NK}   &  0.208 & 0.405 & 0.586 &      0.757 &   7.569 &     0.258 \\
\bottomrule
\end{tabular}
\caption{Zero-shot transfer to the DIODE Outdoor dataset~\cite{diode_dataset}. Best results are in bold, second best are underlined.}
\label{tab:diode-outdoor-full}
\end{table}

\subsection{ZoeDepth with different backbones}

We achieve the best performance for ZoeDepth when using the large BEiT\textsubscript{384}-L \cite{DBLP:journals/corr/abs-2106-08254} backbone for the MiDaS encoder (see Fig.~2 of the main paper), which is responsible for the feature extraction of relative depth computation. According to \Cref{tab:parameters}, this transformer backbone causes ZoeDepth to consist of 345M parameters of which 344M parameters are consumed by MiDaS. In MiDaS, the BEiT\textsubscript{384}-L \cite{DBLP:journals/corr/abs-2106-08254} backbone takes up 305M of the 344M parameters. The number of parameters of ZoeDepth is therefore mainly influenced by the chosen MiDaS encoder backbone.

Due to the modularity of our architecture, we can swap out the MiDaS encoder backbone. In \Cref{tab:parameters}, we compare the number of parameters of ZoeDepth for different backbones. The timm \cite{rw2019timm} repository, which provides the backbones, also offers the base BEiT\textsubscript{384}-B transformer. Utilizing this variant reduces the number of parameters of ZoeDepth from 345M to 112M. Another type of transformer with good performance is the hierarchical transformer Swin2 \cite{liu2022swin} based on shifted windows. When using the base and tiny variants Swin2-B and Swin2-T, the number of parameters of ZoeDepth drops to 102M and 42M, respectively. We report the results of all the aforementioned models evaluated on NYU Depth V2 in \Cref{tab:nyu-backbones}.

\begin{table}[!htb]
\centering
\setlength{\tabcolsep}{9pt} 
\begin{tabular}{llr}
\toprule
\textbf{Method} & \textbf{Encoder}    & \textbf{\# Params} \\
\midrule
Eigen~\etal~\cite{Eigen2014}           & -               & 141M \\
Laina~\etal~\cite{Laina2016}           & ResNet-50       & 64M  \\
Hao~\etal~\cite{Hao2018DetailPD}       & ResNet-101      & 60M  \\
Lee~\etal~\cite{Lee2011}               & -               & 119M \\
Fu~\etal~\cite{Fu2018DeepOR}           & ResNet-101      & 110M \\
SharpNet~\cite{Ramamonjisoa_2019_ICCV} & -               & -    \\
Hu~\etal~\cite{Hu2018RevisitingSI}     & SENet-154       & 157M \\
Chen~\etal~\cite{ijcai2019-98}         & SENet           & 210M \\
Yin~\etal~\cite{Yin_2019_ICCV}         & ResNeXt-101     & 114M \\
BTS~\cite{bts_lee2019big}              & DenseNet-161    & 47M  \\
AdaBins~\cite{bhat2021adabins}         & EfficientNet-B5 & 78M  \\
{LocalBins~\cite{bhat2022localbins}}   & EfficientNet-B5 & 74M  \\
{NeWCRFs~\cite{yuan2022new}}           & Swin-L          & 270M \\
\midrule
\textbf{ZoeDepth (S-L)}          & Swin-L      & 212M \\
\textbf{ZoeDepth (S2-T)}         & Swin2-T     & 42M \\
\textbf{ZoeDepth (S2-B)}         & Swin2-B     & 102M \\
\textbf{ZoeDepth (S2-L)}         & Swin2-L     & 214M \\
\textbf{ZoeDepth (B-B)}          & Beit-B      & 112M \\
\textbf{ZoeDepth (B-L)}          & Beit-L      & 345M \\
\bottomrule
\end{tabular}
\caption{Parameter comparison of ZoeDepth models with different backbones and state of the art models. Note that the number of parameters of ZoeDepth only varies with the backbone and is the same for all variants trained on different dataset combinations, \eg, ZoeD-X-N, ZoeD-M12-N and ZoeD-M12-NK, \etc
}
\label{tab:parameters}
\end{table}

\begin{table}[!htb]
\centering
\setlength{\tabcolsep}{2pt} 
\small
\begin{tabular}{@{}lcccccc@{}}
\toprule
Method & $\delta_1\uparrow$ & $\delta_2\uparrow$ & $\delta_3\uparrow$ & REL~$\downarrow$ & RMSE~$\downarrow$ & $\log_{10}\downarrow$\\ \midrule
BTS~\cite{bts_lee2019big}          & 0.885 & 0.978 & 0.994 & 0.110 & 0.392 & 0.047 \\ 
AdaBins~\cite{bhat2021adabins}     & 0.903 & 0.984 & 0.997 & 0.103 & 0.364 & 0.044 \\ 
LocalBins~\cite{bhat2022localbins} & 0.907 & 0.987 & \underline{0.998} & 0.099 & 0.357 & 0.042 \\ 
NeWCRFs~\cite{yuan2022new}         & 0.922 & 0.992 & \underline{0.998} & 0.095 & 0.334 & 0.041 \\ \midrule
\textbf{ZoeD-M12-N (S-L)}    & 0.937  & 0.992 & \underline{0.998} & 0.086 &  0.310 & 0.037   \\ 
\textbf{ZoeD-M12-N (S2-T)}     & 0.899     & 0.982     & 0.995     & 0.106     & 0.371     & 0.045    \\ 
\textbf{ZoeD-M12-N (S2-B)}    & 0.927     & 0.992     & \textbf{0.999}     & 0.090     & 0.313     & 0.038    \\ 
\textbf{ZoeD-M12-N (S2-L)}    & \underline{0.943}     & \underline{0.993}     & \textbf{0.999}    & \underline{0.083}     & \underline{0.296}     & \underline{0.035}   \\ 
\textbf{ZoeD-M12-N (B-B)}       & 0.922 & 0.990 &  \underline{0.998} & 0.093 & 0.329  & 0.040    \\ 
\textbf{ZoeD-M12-N (B-L)}      & \textbf{0.955} &        \textbf{0.995} &        \textbf{0.999} &             \textbf{0.075} &           \textbf{0.270} &            \textbf{0.032} \\
\bottomrule
\end{tabular}
\caption{Results on the NYU Depth V2 dataset with different backbones. Best results are in bold, second best are underlined.}
\label{tab:nyu-backbones}
\end{table}

\clearpage

\end{document}